\documentclass[preprint,12pt]{elsarticle}
\pdfoutput=1

\usepackage{amssymb}
\usepackage{amsmath}
\usepackage{bm}
\usepackage{makecell}
\usepackage{graphicx}
\usepackage{float}
\usepackage{placeins}
\usepackage{arydshln}
\usepackage{multirow}
\usepackage{csquotes}

\usepackage{natbib}
\usepackage[switch]{lineno}
\usepackage{svg}

\usepackage{siunitx}

\usepackage[%
  breaklinks=true,
  letterpaper=true,
  colorlinks,
  bookmarks=false,
]{hyperref}
\hypersetup{linkcolor={black}, urlcolor={blue}, citecolor={black}, colorlinks=true}

\usepackage[%
  xindy,
  acronym,
]{glossaries}
\glsdisablehyper

\newacronym{gl:DSM}{DSM}{digital surface model}
\newacronym{gl:LoD}{LoD}{levels of detail}
\newacronym{gl:CNN}{CNN}{convolutional neural network}
\newacronym{gl:SSL}{SSL}{self-supervised learning}
\newacronym{gl:MAE}{MAE}{masked autoencoder}
\newacronym{gl:SimMIM}{SimMIM}{simple masked image modeling}
\newacronym{gl:mAP}{mAP}{mean average precision}

\newglossaryentry{stateoftheart}{
  name={state-of-the-art},
  description={the most advanced and developed technology or method at a particular time}
}

\newcommand{\HiResFusedMIMdot}{HiRes-FusedMIM}
\newcommand{\HiResFusedMIM}{\HiResFusedMIMdot\space}

\newcommand{\SimMIMdot}{SimMIM}
\newcommand{\SimMIM}{\SimMIMdot\space}

\usepackage{cleveref}
\usepackage{blindtext}

\journal{ISPRS Journal}

\begin{document}

\begin{frontmatter}



\title{\HiResFusedMIMdot: A High-Resolution RGB-\gls{gl:DSM} Pre-trained Model for Building-Level Remote Sensing Applications}


\author[inst1]{Guneet Mutreja}
\author[inst1]{Philipp Schuegraf}
\author[inst1]{Ksenia Bittner}

\affiliation[inst1]{organization={German Aerospace Center (DLR)},%Department and Organization
            addressline={Münchener Straße 20}, 
            city={Weßling},
            postcode={82234}, 
            state={Bavaria},
            country={Germany}}

%\author[inst2]{Ksenia Bittner}
%\author[inst1,inst2]{Author Three}

\begin{abstract}
% With the recent developments in the generative self-supervised learning domain, foundational models have undoubtedly delivered \gls{stateoftheart} performance for several tasks across multiple domains. However, \glspl{gl:DSM} which are a crucial component to understand urban setting for digital twins, are not yet explored to their full potential from the perspective of a foundational model. Against this background, the research question here is what benefit a foundational model, trained using \glspl{gl:DSM} and conventional RGBs can have on the downstream tasks related to urban scene modelling. We have utilized the power of \gls{gl:SimMIM} to train a foundational model with two encoders: each for an RGB sample and its corresponding \gls{gl:DSM} sample. We observed that adding \glspl{gl:DSM} as an additional input, helped enhance the performances for tasks like building footprints extraction, building height estimation, roof type classification. We found the model trained with separate encoders for each modality works better than a shared encoder in most of the cases. Also, the concatenated features from the two encoders helped both of them to learn from each other.  

Recent advances in self-supervised learning have led to the development of foundation models that have significantly advanced performance in various computer vision tasks. However, despite their potential, these models often overlook the crucial role of high-resolution \glspl{gl:DSM} in understanding urban environments, particularly for building-level analysis, which is essential for applications like digital twins. To address this gap, we introduce \HiResFusedMIMdot, a novel pre-trained model specifically designed to leverage the rich information contained within high-resolution RGB and \gls{gl:DSM} data. \HiResFusedMIM utilizes a dual-encoder \gls{gl:SimMIM} architecture with a multi-objective loss function that combines reconstruction and contrastive objectives, enabling it to learn powerful, joint representations from both modalities. We conducted a comprehensive evaluation of \HiResFusedMIM on a diverse set of downstream tasks, including classification, semantic segmentation, and instance segmentation. Our results demonstrate that: 1) \HiResFusedMIM outperforms previous \gls{stateoftheart} geospatial methods on several building-related datasets, including WHU Aerial and LoveDA, demonstrating its effectiveness in capturing and leveraging fine-grained building information; 2) Incorporating \glspl{gl:DSM} during pre-training consistently improves performance compared to using RGB data alone, highlighting the value of elevation information for building-level analysis; 3) The dual-encoder architecture of \HiResFusedMIMdot, with separate encoders for RGB and \gls{gl:DSM} data, significantly outperforms a single-encoder model on the Vaihingen segmentation task, indicating the benefits of learning specialized representations for each modality. To facilitate further research and applications in this direction, we will publicly release the trained model weights.

\end{abstract}


\begin{highlights}
\item A unique dataset of over 368k paired RGB-\gls{gl:DSM} images at 0.2-0.5 m resolution is curated from various sources
across Europe to train a multimodal pre-trained model- \HiResFusedMIMdot.
\item A novel dual encoder \SimMIM architecture facilitating joint representation learning is developed.
\item Demonstrated the effectiveness of \HiResFusedMIM on a diverse set of building-related tasks, including classification, semantic segmentation, and instance segmentation.
\end{highlights}


\end{frontmatter}


\glsresetall

\section{Introduction }
\label{sec:introduction}

Accurate and up-to-date information about buildings is crucial for a wide range of applications, from urban planning and disaster response to the creation of digital twins. Detailed 3D building models, in particular, are essential for simulating urban environments, assessing risks, and optimizing infrastructure management.

High-resolution imagery, particularly the fusion of RGB and \glspl{gl:DSM}, plays a crucial role in extracting detailed building information. While RGB data captures spectral characteristics of the surface of the Earth, \glspl{gl:DSM} provide critical elevation information, enabling the derivation of building heights and 3D structures. In recent years, foundation models, pre-trained on massive datasets using self-supervised learning, have revolutionized computer vision and shown great promise in remote sensing \cite{satmae,ringmo,bfm,prithvi}. These models learn general-purpose visual representations that can be adapted to various downstream tasks, including those requiring precise building information.

Despite the advantages of foundation models and the importance of high-resolution \glspl{gl:DSM}, existing approaches in remote sensing often fall short in fully leveraging the potential of this combination. This gap stems primarily from the limited availability of large-scale, high-resolution RGB-\gls{gl:DSM} datasets and the inherent challenges in effectively fusing these modalities during pre-training. Models predominantly trained on open-source data like Sentinel-2 imagery (\SI{10}{\meter} resolution) \cite{prithvi} often struggle to generalize well to tasks requiring finer details \cite{cmc}. While some efforts have incorporated high-resolution RGB data \cite{gfm}, the development of pre-trained models that effectively leverage high-resolution \glspl{gl:DSM}, especially for building-level analysis, remains a significant area for explorations.

This paper addresses this gap by introducing \textsc{\HiResFusedMIMdot}, a novel pre-trained model specifically designed for building-level remote sensing applications. \HiResFusedMIM leverages a unique and extensive dataset of paired RGB-\gls{gl:DSM} imagery at \SI{0.2}{}-\SI{0.5}{\meter} resolution, carefully curated from various sources. The model utilizes a dual encoder \gls{gl:SimMIM} architecture \cite{simmim} with a multi-objective loss function that combines both reconstruction and contrastive objectives. We demonstrate the effectiveness of this approach on a diverse range of downstream tasks, including classification, semantic segmentation, and instance segmentation, showing its superior performance in capturing and leveraging high-resolution, multi-modal information for detailed building analysis. 

The paper is organized as follows: Section \ref{sec:relatedwork} reviews related work on importance of \glspl{gl:DSM}, self-supervised learning, and foundation models in remote sensing. Section \ref{sec:methodology} details the \HiResFusedMIM methodology, including the dataset, model architecture, and pre-training process. Section \ref{sec:experiments} presents the experimental results, comparing our approach to relevant baselines and \gls{stateoftheart} methods. Section \ref{sec:ablationstudies} presents a series of ablation studies analyzing the impact of different model components. Finally, Section \ref{sec:Conclusion} discusses the implications of our findings and outlines future research directions.

\section{Related work}
\label{sec:relatedwork}

This section provides a comprehensive review of existing research relevant to the development and applications of \HiResFusedMIMdot. We first discuss the significance of \glspl{gl:DSM} in remote sensing, followed by a discussion of recent advancements in self-supervised learning and foundation models for remote sensing. Finally, we highlight the motivation and contributions of our research, emphasizing the novelty of our approach.

\subsection{The Importance of \glspl{gl:DSM} in Remote Sensing} 
\Glspl{gl:DSM} are essential data sources in remote sensing, providing crucial elevation information about the surface of the Earth and all objects upon it. Unlike digital terrain models (DTMs), which represent the bare earth, DSMs capture the heights of buildings, vegetation, trees, and other structures. This detailed elevation data is essential for a wide range of applications, including:
\begin{itemize}
\item Urban Planning and Analysis: \glspl{gl:DSM} are used to analyze urban morphology, model urban growth, assess urban density, and aid in infrastructure planning and management.
\item Forestry and Vegetation Management: \glspl{gl:DSM} are used to estimate forest canopy heights, analyze forest structure, monitor deforestation, and assess biomass.
\item Disaster Management: \glspl{gl:DSM} play a vital role in flood modeling, landslide risk assessment, and damage assessment after natural disasters.
\item 3D Building Reconstruction: \glspl{gl:DSM} provide the height information necessary to create realistic 3D building models, which are essential for applications like urban planning, virtual city modeling, and digital twins.
\end{itemize}

The recent advancements in deep learning have significantly improved the utilization of \glspl{gl:DSM} in remote sensing, enabling the development of more automated and accurate methods for various tasks. Deep learning models, particularly \glspl{gl:CNN}, have demonstrated success in:

\begin{itemize}
\item Semantic Segmentation of \glspl{gl:DSM}: \glspl{gl:CNN} can effectively classify \gls{gl:DSM} pixels into different categories, such as buildings, vegetation, and water bodies, enabling the automatic extraction of features with high accuracy \cite{extraction, fusion, philipp, philipp2}.
\item Joint Learning from RGB and \gls{gl:DSM} Data: By combining RGB and \gls{gl:DSM} data, deep learning models can learn richer representations, leveraging both spectral and elevation information  to achieve superior performance in tasks like land cover classification, object detection, and semantic segmentation \cite{fusion, autolod2, philipp3, planes4lod2}.
\item Generating \gls{gl:LoD}2-Level Building Models: Recent studies have demonstrated the potential of deep learning in achieving \gls{gl:LoD}2-level 3D building reconstruction by extracting roof planes and other detailed features from \glspl{gl:DSM}  \cite{autolod2, lod2.2, planes4lod2, philipp4}.
\end{itemize}

This progress in deep learning, combined with the increasing availability of high-resolution \gls{gl:DSM} data,  has opened new possibilities for extracting valuable information from remote sensing data and addressing complex challenges in various domains.

\subsection{Self-Supervised Learning and Foundation Models in Remote Sensing} The field of remote sensing has been significantly impacted by the recent advancements in \gls{gl:SSL} and foundation models. \Gls{gl:SSL} enables models to learn valuable representations from unlabeled data, reducing the reliance on costly and often limited labeled datasets \cite{ssl}. This capability is particularly beneficial in remote sensing, where obtaining large-scale, high-quality labeled data is challenging.

Foundation models, typically large-scale neural networks pre-trained on massive datasets using~\gls{gl:SSL} objectives, have emerged as a powerful paradigm for various computer vision tasks. These models learn general-purpose visual representations that can be effectively adapted to a wide range of downstream tasks, often achieving~\gls{stateoftheart} performance with minimal fine-tuning.

% In remote sensing, foundation models have demonstrated significant promise in tasks like scene classification, object detection, and semantic segmentation. Models such as:
% \begin{itemize}
%     \item SatMAE \cite{satmae}: A transformer-based model pre-trained on temporal and multi-spectral satellite imagery using a \gls{gl:MAE} approach.
%     \item RingMo \cite{ringmo}: An autoencoder-based model pre-trained on a large collection of very high-resolution RGB images, specifically designed for dense, small object detection.
%     \item GFM \cite{gfm}: A geospatial foundation model that uses a continual pre-training paradigm and a multi-objective loss function to learn robust representations.
%     \item Prithvi \cite{prithvi}: A transformer-based model pre-trained on a large scale Harmonized Landsat-Sentinel 2 (HLS) dataset, demonstrating applicability to real-world tasks like flood mapping and crop segmentation.
%     \item The Billion-Scale Foundation Model \cite{bfm}: A study exploring the scaling of model parameters in remote sensing foundation models, showing promising results on tasks like rotated object detection and semantic segmentation.
% \end{itemize}

In remote sensing, foundation models have demonstrated significant promise in tasks such as scene classification, object detection, and semantic segmentation. For instance, ~\citet{satmae} developed a transformer-based model pre-trained on temporal and multi-spectral satellite imagery using a~\gls{gl:MAE} approach. This model effectively captures temporal and spectral information, enhancing performance in downstream tasks. Similarly, ~\citet{ringmo} utilizes an autoencoder architecture pre-trained on a large collection of very high-resolution RGB images, specifically designed to address the challenges of dense, small object detection in remote sensing data.

Another notable example is the work by \citet{gfm}, which employs a continual pre-training paradigm and a multi-objective loss function to learn robust representations. \citet{prithvi} proposed a transformer-based model pre-trained on a large-scale Harmonized Landsat-Sentinel 2 (HLS) dataset, demonstrating its effectiveness in real-world applications like flood mapping and crop segmentation.

Furthermore, \citet{bfm} explored the scaling of model parameters in remote sensing foundation models. The research shows promising results on tasks such as rotated object detection and semantic segmentation, indicating that scaling up model size can capture more complex patterns inherent in remote sensing data.

These models demonstrate the power of \gls{gl:SSL} pre-training in capturing meaningful patterns and features from remote sensing data. However, a critical gap remains in fully leveraging the potential of high-resolution \gls{gl:DSM} data within this pre-training paradigm, especially for tasks focused on detailed building-level analysis.

\subsection{Bridging the Gap: Towards High-Resolution, Multi-Modal Foundation Models for Building Analysis} Despite the promising advancements in foundation models, their application to building-level remote sensing tasks remains limited. Existing models primarily rely on medium-resolution satellite data like Sentinel-2 \cite{satmae, prithvi, satlas} or focus solely on RGB information \cite{gfm, scalemae}, neglecting the valuable elevation data provided by high-resolution \glspl{gl:DSM}.

While some multi-modal models have been proposed, they haven't effectively addressed the challenges of high-resolution \glspl{gl:DSM} and their potential for building analysis. For example:

\begin{itemize}
    \item DeCUR \cite{decur}: While demonstrating promising results on RGB-DEM fusion for semantic segmentation, their experiments used a smaller dataset data compared to the large-scale data employed in our research.
    \item MultiMAE \cite{multimae}: This approach extends the MAE framework to accept multiple modalities, but its performance when integrating high-resolution \gls{gl:DSM} data with RGB remains unexplored, especially for building-related tasks.
\end{itemize}

Although other multi-modal models have been pre-trained, such as \citet{msgfm} and \citet{skysense}, they do not explicitly emphasize the use of \glspl{gl:DSM} for building-level tasks. This underutilization of high-resolution \glspl{gl:DSM} persists despite their proven value in extracting building heights, reconstructing roof shapes, and generating detailed 3D building models \cite{autolod2, 3Drecons}.

Our research seeks to address this gap by developing \HiResFusedMIMdot, a pre-trained model explicitly designed to leverage high-resolution RGB-\gls{gl:DSM} data for building-level analysis. We will demonstrate how effectively integrating these modalities during pre-training leads to superior performance in downstream tasks that require fine-grained building information.

\subsection{Contributions} This paper addresses this critical gap of underutlization of \glspl{gl:DSM} by introducing \HiResFusedMIMdot, a novel pre-trained model designed explicitly to leverage high-resolution RGB-\gls{gl:DSM} data for pre-training, with a specific focus on building-related remote sensing applications. Our key contributions include:

\begin{itemize}
    \item Curating a Large-Scale, High-Resolution, Paired RGB-\gls{gl:DSM} Dataset: We curated a unique dataset of over 368k paired RGB and \gls{gl:DSM} images at  \SI{0.2}{}-\SI{0.5}{\meter} resolution, meticulously collected from various sources across Europe. This dataset, while not publicly shared due to licensing restrictions, serves as a vital foundation for training our model and highlights the need for more publicly available high-resolution, multi-modal datasets in the remote sensing community.
    \item Effective Multi-Modal Fusion with a Dual Encoder \SimMIM Architecture: \HiResFusedMIM utilizes a dual encoder \SimMIM architecture \cite{simmim} with separate encoders for RGB and \gls{gl:DSM} data. This architecture facilitates efficient joint representation learning, capturing both modality-specific and cross-modal features. A contrastive loss (InfoNCE) \cite{infonce} is introduced between the encoder outputs to explicitly encourage the alignment of RGB and \gls{gl:DSM} representations, further enhancing the ability of the model to learn from both modalities.
    \item Superior Performance on Building-Level Downstream Tasks: We demonstrate the effectiveness of \HiResFusedMIM on a diverse set of building-related tasks, including classification, semantic segmentation, and instance segmentation. Our experiments showcase superior performance compared to models trained on lower-resolution or single-modality data, validating the benefits of incorporating high-resolution \glspl{gl:DSM} in the pre-training process. To facilitate further research in this direction, we will publicly share the trained model weights.
\end{itemize}

\section{Methodology}
\label{sec:methodology}

\subsection{Pre-Training Dataset – A High-Resolution RGB-\gls{gl:DSM} Collection}
A critical component of \HiResFusedMIM is the creation of a unique, large-scale dataset of paired RGB and \gls{gl:DSM} imagery, carefully curated to facilitate the learning of representations optimized for building-level analysis. Addressing the limited availability of publicly accessible high-resolution \gls{gl:DSM} data, we combined multiple sources to construct this dataset. A major chunk of the data was sourced from open-source geoportals provided by local governments across Europe. We targeted cities that offer aerial imagery and corresponding \glspl{gl:DSM} at high resolutions (\SI{20}{\centi\metre} - \SI{50}{\centi\metre}), striving to encompass a wide variety of building types, urban layouts, and geographical variations. \Cref{tab:datasourcesaerial} and \Cref{tab:datasourcessatellite} list the specific city sources, spatial resolutions, and the number of image pairs acquired from public geoportals and commercial providers, respectively.

% \usepackage{graphicx}
\begin{table*}[!ht]
\centering
\caption{ Aerial Imagery and \gls{gl:DSM} Data Sources (Public Geoportals).\label{tab:datasourcesaerial}}\vspace{0.2cm}
\small
\begin{tabular}{ccccc}
\hline
Country  & City/Region & \makecell{Spatial Resolution \\(in m)} & \makecell{Number of \\DSM-RGB Pairs} \\ \hline
Germany      & Augsburg       & 0.5       & 26928   \\
Germany      & Berlin      &  0.5        & 24186     \\
Germany      & Bochum       &  0.5        & 1936     \\
Switzerland      & Appenzel       &  0.5        & 289     \\
Switzerland      & Basel       & 0.5       & 2300   \\
Switzerland      & Bern       & 0.5       & 6794   \\
Switzerland      & BuchSG       &  0.5        & 2288     \\
Switzerland      & Geneva       & 0.5       & 8108  \\
Switzerland      & Herisau       & 0.5       & 884   \\
Switzerland      & Lausanne       &  0.5        & 5355     \\
Switzerland      & St. Gallen       & 0.5       & 3111   \\
Switzerland      & Winterthur       & 0.5       & 1352   \\
Switzerland      & Zurich       & 0.5       & 3172   \\ \hline

\end{tabular}%
\end{table*}

\begin{table*}[!ht]
\centering
\caption{ Satellite Imagery and \gls{gl:DSM} Data Sources (Commercial and Other Sources).\label{tab:datasourcessatellite}}\vspace{0.2cm}
\small
\begin{tabular}{ccccc}
\hline
Country  & City/Region & \makecell{Spatial Resolution \\(in m)} & \makecell{Number of \\DSM-RGB Pairs} \\ \hline
Brazil       & São Paulo      &  0.5        & 19320     \\
Colombia     & Medellin       &  0.3        & 1539     \\
Germany      & Berlin      &  0.3        & 21164     \\
Germany      & Dresden       & 0.2       & 47526   \\
Germany      & Hamburg      &  0.5        & 1734     \\
Germany      & Juelich       &  0.5        & 8066     \\
Germany      & Munich       & 0.5       & 8712   \\
Germany      & North Rhine-Westphalia       & 0.3-0.5       & 131224   \\
Libya      & Tripoli      &  0.5        & 16146     \\
United States      & Washington       &  0.5        & 2793     \\
Vietnam      & Ho Chi Minh City       & 0.5       & 143   \\
Jordan      & Jordan       & 0.5       & 846   \\
Kyrgyzstan      & Kyrgyzstan       & 0.5       & 15240   \\\hline

\end{tabular}%

\end{table*}

\paragraph{DSM Data Underutilization: A Critical Gap} Despite the clear advantages of \glspl{gl:DSM} for detailed building analysis, their use in pre-training remote sensing models remains surprisingly limited.  \Cref{tab:dsmusage} presents a comparative analysis of \gls{gl:DSM} data utilization in various pre-trained remote sensing models. As the table highlights, most existing models either do not incorporate \glspl{gl:DSM} or use them in limited quantities, often relying on datasets where \glspl{gl:DSM} are not the primary focus. This underutilization persists despite evidence suggesting that \glspl{gl:DSM} can significantly improve performance in tasks such as building segmentation and 3D modeling.

\begin{table*}[t]
\centering
\caption{DSM Data Utilization in Pre-trained Remote Sensing Models.\label{tab:dsmusage}}\vspace{0.2cm}
\small
\begin{tabular}{ccc}
\hline
Model    &\makecell{DSM Data Included \\ in Pre-training} & \makecell{Approximate Number \\ of \gls{gl:DSM} Images in \\ Pre-training Dataset} \\ \hline
SatMAE  \cite{satmae}               & No      & 0     \\
RingMo  \cite{ringmo}  & No      & 0      \\
GFM   \cite{gfm}  & No      & 0         \\
DeCUR  \cite{decur}      & Yes      & 6942     \\
SatlasPretrain   \cite{satlas}    & No      & 0      \\
CROMA     \cite{croma}      & No      & 0      \\
msGFM      \cite{msgfm}      & Yes      & 40k         \\
Scale-MAE       \cite{scalemae}      & No      & 0         \\\hline
\HiResFusedMIM (Ours)                    & Yes      & ~368k         \\\hline

\end{tabular}%

\end{table*}

The limited use of \glspl{gl:DSM} in pre-training underscores the importance and novelty of our approach. By creating a large-scale, high-resolution dataset specifically tailored for RGB-\gls{gl:DSM} fusion and using it to pre-train \HiResFusedMIMdot, our research directly addresses this gap, enabling the development of models that can effectively leverage the rich information contained within \glspl{gl:DSM} for building-related analysis.

\subsection{\HiResFusedMIM Model Architecture}

\begin{figure*}[t]
    \tabcolsep=0.04cm
	\centering
	% \includesvg[width=\textwidth]{figures/Methodology.sv{"originatingScript":"m2","payload":{"guid":"d7147572-bbf5-4d28-8875-71719d0b07c7f34ed3","muid":"433374d3-2efc-4168-89a4-64da4fe1f1c6f250fa","sid":"3ff91db7-d878-4266-ba8c-273097266e99cfc4c2"}}g}
    \includegraphics[width=\textwidth]{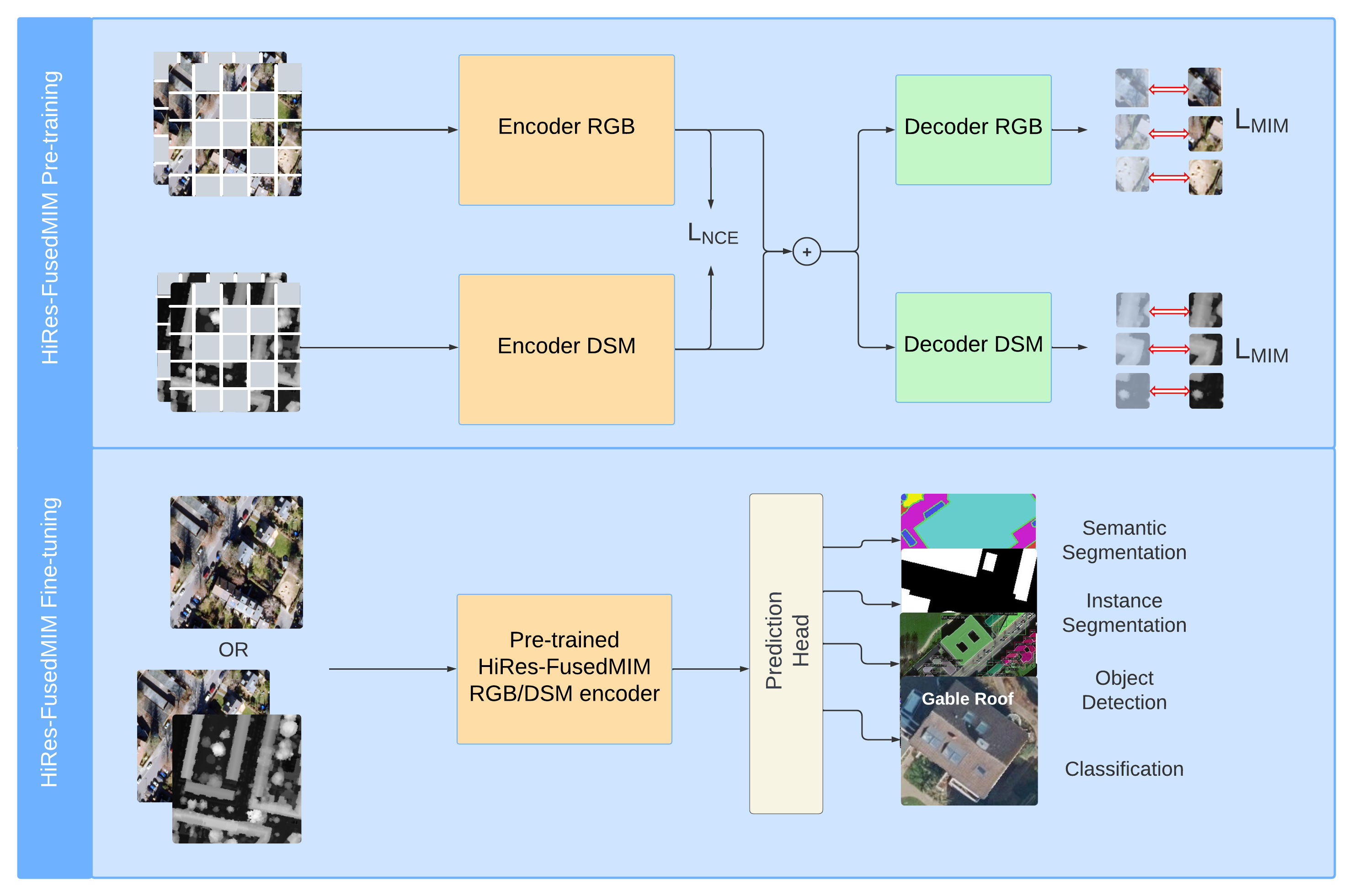}
	\caption{HiRes-Fused pre-training and fine-tuning.\label{fig:arch}}
	\vspace{-0.2cm}
\end{figure*}

\HiResFusedMIM is designed to learn joint representations from paired high-resolution RGB and \gls{gl:DSM} data, employing a dual encoder architecture based on the \gls{gl:SimMIM} framework \cite{simmim}. \Cref{fig:arch} provides a detailed illustration of the architecture.

\paragraph{Dual Encoders and Decoders} Each encoder branch processes its respective modality—the RGB encoder operates on 3-channel RGB images, while the \gls{gl:DSM} encoder takes a single-channel \gls{gl:DSM}. Both encoders are implemented using a Swin transformer architecture \cite{swin} to effectively capture local and global image context, which is especially crucial for high-resolution imagery. The model includes two separate decoder branches, one for reconstructing the masked RGB patches and another for reconstructing the masked \gls{gl:DSM} patches. Each decoder is a lightweight network, designed to map the encoded features back to the original image space.

\paragraph{Feature Concatenation} To facilitate cross-modal interaction and the learning of joint representations, the features from the final layer of each encoder are concatenated channel-wise, resulting in a feature vector of dimension of $2048$. This concatenated feature vector is then passed to both decoders.

\paragraph{Reconstruction Loss} The decoders are trained to reconstruct the masked patches from their respective modalities, supervised by a reconstruction loss. We employ the L1 loss \cite{simmim} for this purpose, defined as:

\begin{align}
\label{eq:L1loss}
\mathcal{L}_{\text{MIM}} = \frac{||\bm{X}_{\text{RGB}} - \bm{R}_{\text{RGB}}||_1 + ||\bm{X}_{\text{DSM}} - \bm{R}_{\text{DSM}}||_1}{N},
\end{align}
where $\bm{X}_{\text{RGB}}$ and $\bm{X}_{\text{DSM}}$ are original pixel values of masked RGB and \gls{gl:DSM} patches, respectively.
$\bm{R}_{\text{RGB}}$ and $\bm{R}_{\text{DSM}}$ are reconstructed pixel values (model outputs) for the RGB and \gls{gl:DSM} patches, respectively.
$N$ is the total number of masked pixels across both RGB and \gls{gl:DSM} patches.

\paragraph{Contrastive Loss (InfoNCE)} To further encourage the alignment of RGB and \gls{gl:DSM} representations, we introduce a contrastive loss between the output embeddings of the two encoders. We utilize the InfoNCE loss \cite{infonce}, which encourages similar representations for positive pairs (RGB and \gls{gl:DSM} embeddings from the same location) while pushing dissimilar representations for negative pairs (embeddings from different locations). The InfoNCE loss is defined as:
\begin{align}
\label{eq:ContrastiveLoss}
\mathcal{L}_{\text{InfoNCE}} = - \log \left( \frac{\exp(\text{sim}(\bm{z}_{i}, \bm{z}_{j}) / \tau)}{\sum_{k=1}^{K} \exp(\text{sim}(\bm{z}_{i}, \bm{z}_{k}) / \tau)} \right)
\end{align}
where, 
$\bm{z}_i$ and $\bm{z}_j$ are the output embeddings from RGB and \gls{gl:DSM} encoders for a positive pair. $\tau$ represents temperature parameter, $K$ represents the number of negative samples and $\text{sim}(\bm{z}_i, \bm{z}_j)$ is the Cosine similarity between embeddings $\bm{z}_i$ and $\bm{z}_j$.

\paragraph{Multi-Objective Loss Function} The total loss function used during pre-training combines the reconstruction losses and the contrastive loss with a weighting factor ($\alpha$) to balance their contributions:

\begin{align}
\label{eq:TotalLoss}
\mathcal{L}_{\text{total}} = (1 - \alpha) * \mathcal{L}_{\text{MIM}} + \alpha * \mathcal{L}_{\text{InfoNCE}}
\end{align}
We set $\alpha$ = 0.05, allocating 95\% of the weight to the MIM objective and 5\% to the contrastive loss. This weighting encourages the model to prioritize accurate reconstruction while benefiting from the representation alignment provided by the contrastive loss.

\paragraph{Pre-Training Details} Before pre-training, we performed per-city normalization on the RGB and \gls{gl:DSM} images. For each city represented in our dataset, we calculated the mean ($\mu$) and standard deviation ($\sigma$) values across all RGB channels and \gls{gl:DSM} channel. Each image (I) from that city was then normalized using the following equation:

\begin{align}
\label{eq:normalization}
I_{\text{normalized}} = \frac{I - \mu}{\sigma}
\end{align}

This normalization helps mitigate variations in illumination, atmospheric conditions, and sensor characteristics across different sources.

During pre-training, we randomly masked 60\% of the input patches in each training iteration, a strategy consistent with findings that suggest larger masking ratios can lead to improved performance \cite{gfm, simmim}. We augmented the training data with random horizontal flipping and random cropping to encourage the model to learn more robust representations. The AdamW optimizer \cite{adam} was used with a base learning rate of 5e-4 and a cosine learning rate schedule. The model was trained for 400 epochs with a batch size of 128 per GPU on 8 NVIDIA V100 GPUs and an image size of $224\times224$.

\section{Experiments and Results}
\label{sec:experiments}

This section presents a comprehensive evaluation of \HiResFusedMIMdot, highlighting its effectiveness in learning powerful representations from high-resolution RGB-\gls{gl:DSM} data for building-related tasks. We first demonstrate the reconstruction capabilities of the model after pre-training, visually highlighting its ability to capture intricate details from both modalities. We then analyze the performance of the model on various downstream tasks, comparing it to relevant baselines and existing foundation models.

\subsection{Pre-Training Evaluation: Qualitative Analysis of Reconstructions}
\begin{figure*}
\tabcolsep=0.05cm
\centering
\begin{tabular}{cccc}
       {\fontsize{8}{10}\selectfont{Ground Truth}} & {\fontsize{8}{10}\selectfont{Reconstruction}} &{\fontsize{8}{10}\selectfont{Ground Truth}} & {\fontsize{8}{10}\selectfont{Reconstruction}} \\
       \vspace{0.1cm}
       
       \raisebox{-.5\height}{\frame{\includegraphics[width=0.23\textwidth]{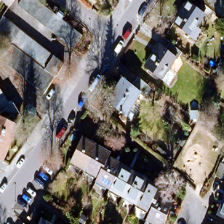}}} &
       \raisebox{-.5\height}{\frame{\includegraphics[width=0.23\textwidth]{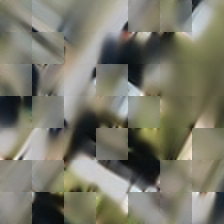}}} &
       \raisebox{-.5\height}{\frame{\includegraphics[width=0.23\textwidth]{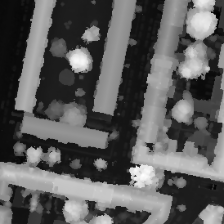}}} &
       \raisebox{-.5\height}{\frame{\includegraphics[width=0.23\textwidth]{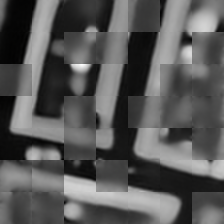}}} \\
       \vspace{0.1cm}
       
       \raisebox{-.5\height}{\frame{\includegraphics[width=0.23\textwidth]{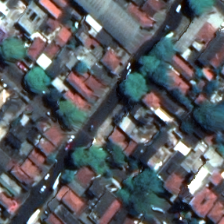}}} &
       \raisebox{-.5\height}{\frame{\includegraphics[width=0.23\textwidth]{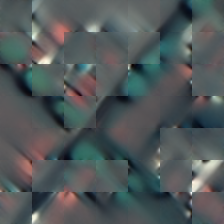}}} &
       \raisebox{-.5\height}{\frame{\includegraphics[width=0.23\textwidth]{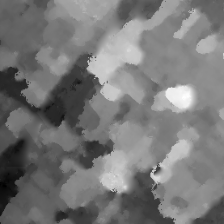}}} &
       \raisebox{-.5\height}{\frame{\includegraphics[width=0.23\textwidth]{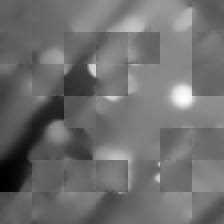}}} \\

       \vspace{0.1cm}
       
       \raisebox{-.5\height}{\frame{\includegraphics[width=0.23\textwidth,height=91pt]{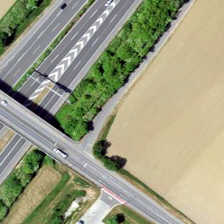}}} &
       \raisebox{-.5\height}{\frame{\includegraphics[width=0.23\textwidth]{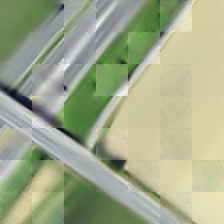}}} &
       \raisebox{-.5\height}{\frame{\includegraphics[width=0.23\textwidth,height=91pt]{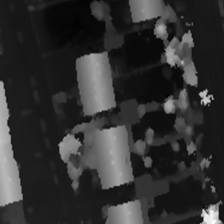}}} &
       \raisebox{-.5\height}{\frame{\includegraphics[width=0.23\textwidth]{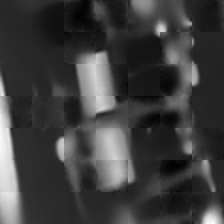}}} \\
       \vspace{0.1cm}
       
       \raisebox{-.5\height}{\frame{\includegraphics[width=0.23\textwidth]{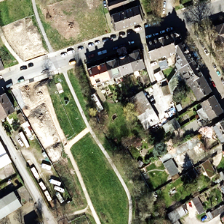}}} &
       \raisebox{-.5\height}{\frame{\includegraphics[width=0.23\textwidth]{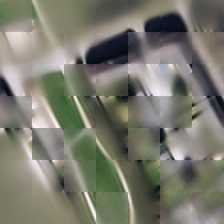}}} &
       \raisebox{-.5\height}{\frame{\includegraphics[width=0.23\textwidth]{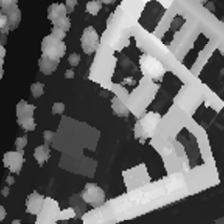}}} &
       \raisebox{-.5\height}{\frame{\includegraphics[width=0.23\textwidth]{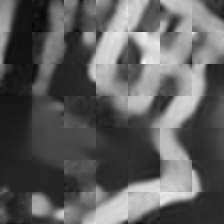}}} \\
       \vspace{0.1cm}
       
       \raisebox{-.5\height}{\frame{\includegraphics[width=0.23\textwidth]{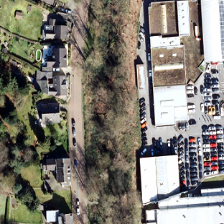}}} &
       \raisebox{-.5\height}{\frame{\includegraphics[width=0.23\textwidth]{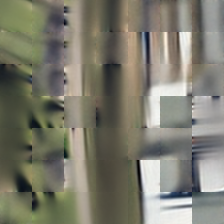}}} &
       \raisebox{-.5\height}{\frame{\includegraphics[width=0.23\textwidth]{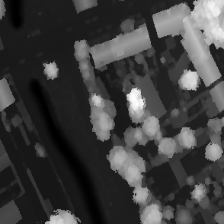}}} &
       \raisebox{-.5\height}{\frame{\includegraphics[width=0.23\textwidth]{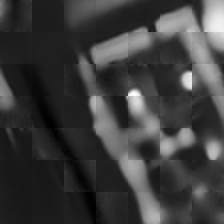}}} \\
\end{tabular}
\caption{Visualization of the reconstruction capabilities of the pretrained model. The left two columns represent ground truth and reconstructions from RGB samples, while the right two columns depict ground truth and reconstructions from \gls{gl:DSM} samples.\label{fig:resvis1}}
\vspace{-0.2cm}
\end{figure*}

\Cref{fig:resvis1} showcases the ability of \HiResFusedMIM to reconstruct masked patches from both RGB and \gls{gl:DSM} inputs after pre-training. The figure presents examples of input images with masked regions, the corresponding reconstructed outputs, and the original unmasked images.

As evident from the visualizations in \Cref{fig:resvis1}, \HiResFusedMIM achieves impressive reconstruction quality for both RGB and \gls{gl:DSM} data. The model accurately recovers fine details, such as building edges, roof structures, and subtle variations in elevation, indicating its ability to learn meaningful representations of the underlying spatial structures and cross-modal relationships. These strong reconstruction capabilities suggest that the pre-trained model has acquired valuable knowledge that can be effectively leveraged for downstream tasks requiring detailed building information.

\subsection{Image Classification}
We evaluated \HiResFusedMIM on two widely used classification benchmarks: the UC Merced Land Use Dataset (UCM) and the BigEarthNet (BEN) dataset. UCM, with its 21 land-use classes and relatively high resolution (0.3m), allows us to assess the performance of the model on data similar in resolution to the pre-training data. BEN, a multi-label classification dataset with 12-band Sentinel-2 imagery at 10m, 20m, and 60m resolutions, enables us to analyze generalization capabilities of the model when it is applied to lower-resolution data. We followed the same classification pipeline as described in \cite{gfm}, utilizing Swin-Base transformer with a linear classifier head.

\begin{table*}[t]
\centering
\caption{ UC Merced classification accuracy and BigEarthNet multi-label 
 classification mean average precision results.\label{tab:classification_results}}\vspace{0.2cm}
\small
\begin{tabular}{llll}
\hline
Method         & UCM & BEN 1\% & BEN 10\%\\ \hline
ResNet50 (ImageNet-1k) \cite{resnet}    & 98.8      & 41.3 & 80.0  \\
SeCo  \cite{seco}    & 97.1  & 63.6 & 82.6      \\
ViT (ImageNet-22k)  \cite{vit}    & 93.1      & 73.6 & 84.7  \\
SatMAE    \cite{satmae}       & 92.6  & 68.9 & 81.8      \\
Swin-B (random)   \cite{swin}   & 66.9      & 65.7 & 80.6  \\
Swin-B (ImageNet-22k)   \cite{swin}    & 99.0  & 79.5 & 85.7      \\
GFM    \cite{gfm}           & \textbf{99.0}      & \textbf{80.7} & \textbf{86.3}  \\ \hline
\HiResFusedMIM   & 98.1  & 75.9 & 84.7      \\ \hline

\end{tabular}%

\end{table*}

\Cref{tab:classification_results} presents the classification accuracy of \HiResFusedMIM on the UCM and BEN datasets. Note that the results for other models were obtained from their respective publications and are provided here  for context and comparison with our proposed approach. We have not reproduced these algorithms at our end. While not achieving \gls{stateoftheart} results, \HiResFusedMIM demonstrates strong performance, achieving 98.1\% accuracy on UCM and 84.7\% \gls{gl:mAP} on BEN (10\% data). These results are particularly encouraging considering the relatively small size of our pre-training dataset (around 368k images) compared to the larger datasets used by top-performing models like GFM (600k images). This suggests that our focus on curating a high-quality, high-resolution, and multi-modal dataset can lead to competitive performance even with a more compact training set.
Furthermore, on the BigEarthNet dataset, which uses lower-resolution Sentinel-2 imagery, \HiResFusedMIM outperforms other foundation models like SeCo and SatMAE, both of which are pre-trained solely on Sentinel-2 data. Our findings align with the observations in \cite{cmc}, where models pre-trained on higher-resolution data exhibited an advantage in classifying lower-resolution imagery. This highlights the potential of \HiResFusedMIM to generalize effectively across different spatial resolutions, making it suitable for a wider range of remote sensing applications.

\subsection{Semantic Segmentation}
To evaluate the effectiveness of \HiResFusedMIM for semantic segmentation, we conducted experiments on five datasets:
\begin{itemize}
\item WHU Aerial Building Dataset \cite{whuaerial}: A high-resolution (\SI{0.3}{\meter} GSD) dataset for building segmentation, where we evaluated the RGB encoder of \HiResFusedMIMdot.
\item LoveDA Dataset \cite{LoveDA}: A domain adaptation dataset (\SI{0.3}{\meter} resolution) featuring seven land cover classes. The dataset encompasses both urban and rural domains, which brings considerable challenges due to the: 1) multi-scale objects; 2) complex background samples; and 3) inconsistent class distributions. 
\item Vaihingen Dataset \cite{vaihingen}: A benchmark dataset with both RGB and \gls{gl:DSM} data (\SI{0.9}{\meter} resolution), enabling assessment of multi-modal learning capabilities. We evaluate both the RGB and the combined RGB-\gls{gl:DSM} encoders of \HiResFusedMIMdot.
\item GeoNRW Dataset \cite{geonrw}: This dataset provides a diverse set of urban scenes, including orthorectified aerial photographs, LiDAR-derived digital elevation models (DEMs), and OpenStreetMap-refined segmentation maps. It covers $10$ classes and features data from the German state of North Rhine-Westphalia. The dataset is particularly relevant for our work, as it allows us to assess the performance of \HiResFusedMIM on both RGB and \gls{gl:DSM} modalities.
\item SpaceNet Building Detection Dataset (V1) \cite{spacenet}: A large-scale dataset with high-resolution (\SI{0.5}{\meter}) imagery for building footprint extraction.
\end{itemize}

\begin{figure*}
\tabcolsep=0.05cm
\centering
\begin{tabular}{ccccc}
        {\fontsize{8}{10}\selectfont{}} & {\fontsize{8}{10}\selectfont{Ground Truth}} & {\fontsize{8}{10}\selectfont{Segmented Mask}} &{\fontsize{8}{10}\selectfont{Ground Truth}} & {\fontsize{8}{10}\selectfont{Segmented Mask}} \\
       \vspace{0.1cm}
       \rotatebox[origin=c]{90}{\fontsize{8}{10}\selectfont{WHU Aerial}} &\raisebox{-.5\height}{\frame{\includegraphics[width=0.23\textwidth]{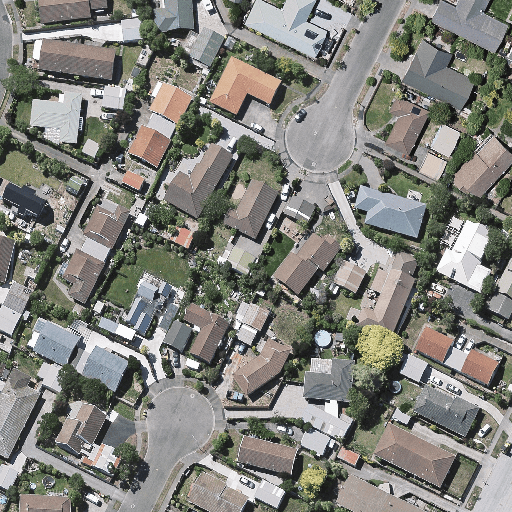}}}&\raisebox{-.5\height}[0pt]{\frame{\includegraphics[width=0.23\textwidth]{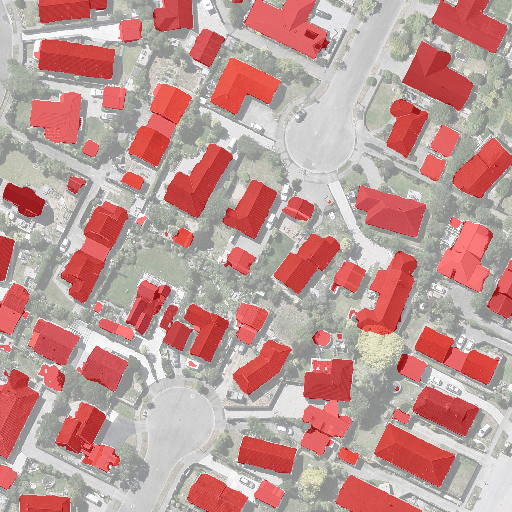}}}&\raisebox{-.5\height}[0pt]{\frame{\includegraphics[width=0.23\textwidth,height=91pt]{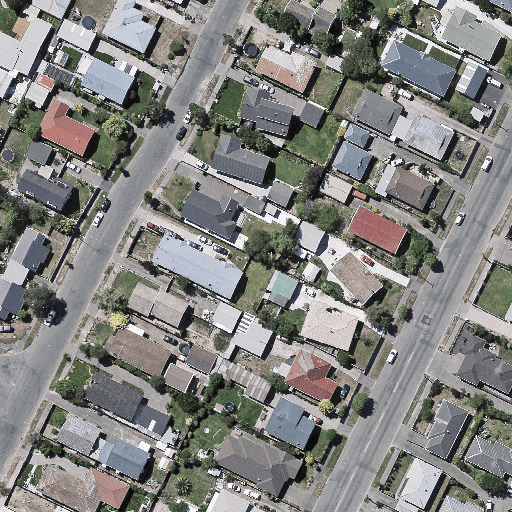}}}&\raisebox{-.5\height}{\frame{\includegraphics[width=0.23\textwidth]{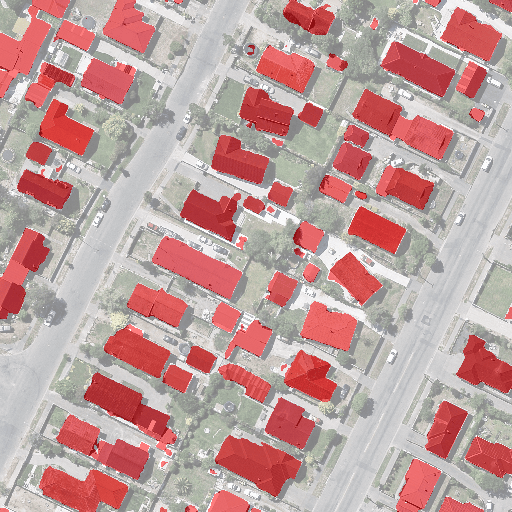}}}\vspace{0 cm}\\\rotatebox[origin=c]{90}{\fontsize{8}{10}\selectfont{LoveDA}}&\raisebox{-.5\height}{\frame{\includegraphics[width=0.23\textwidth]{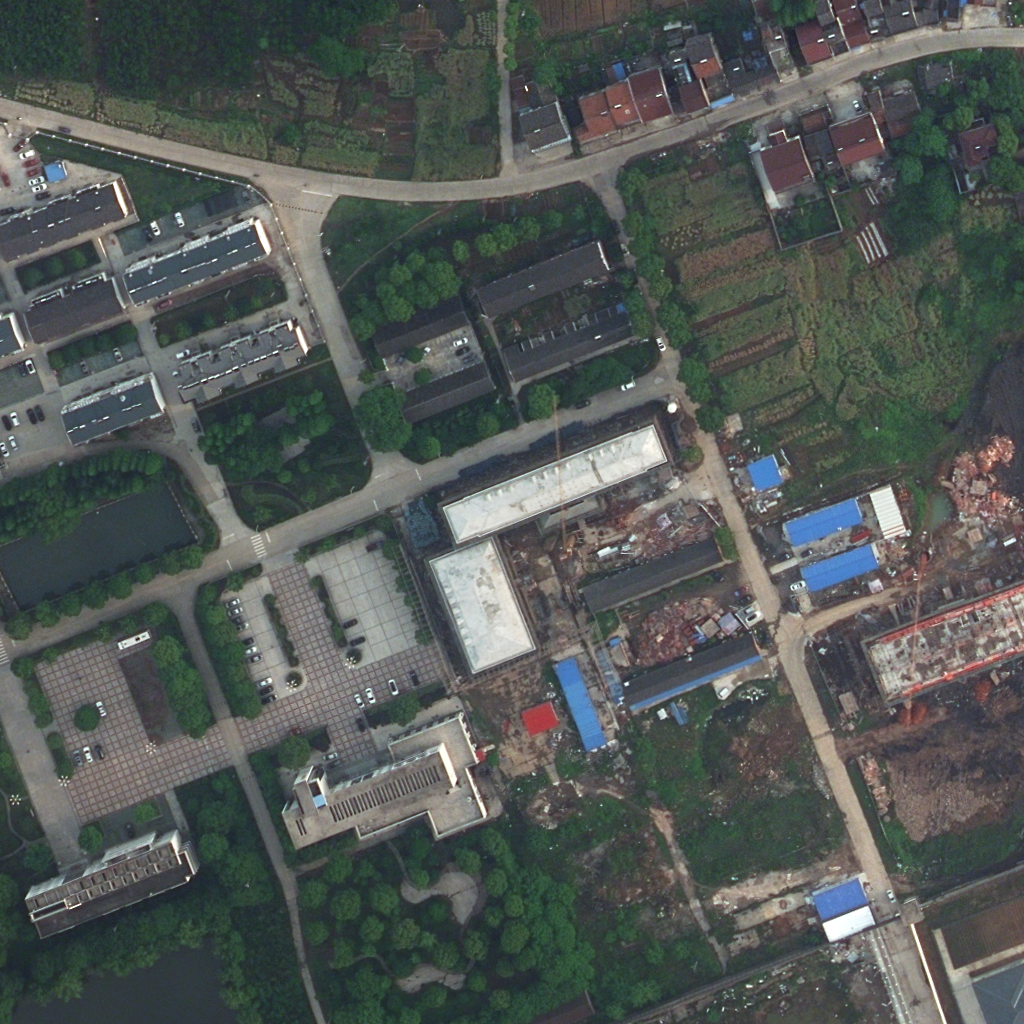}}}&\raisebox{-.5\height}{\frame{\includegraphics[width=0.23\textwidth]{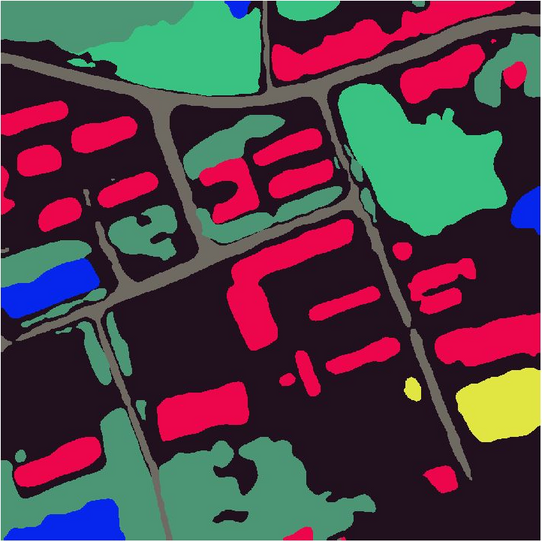}}}&\raisebox{-.5\height}{\frame{\includegraphics[width=0.23\textwidth]
       {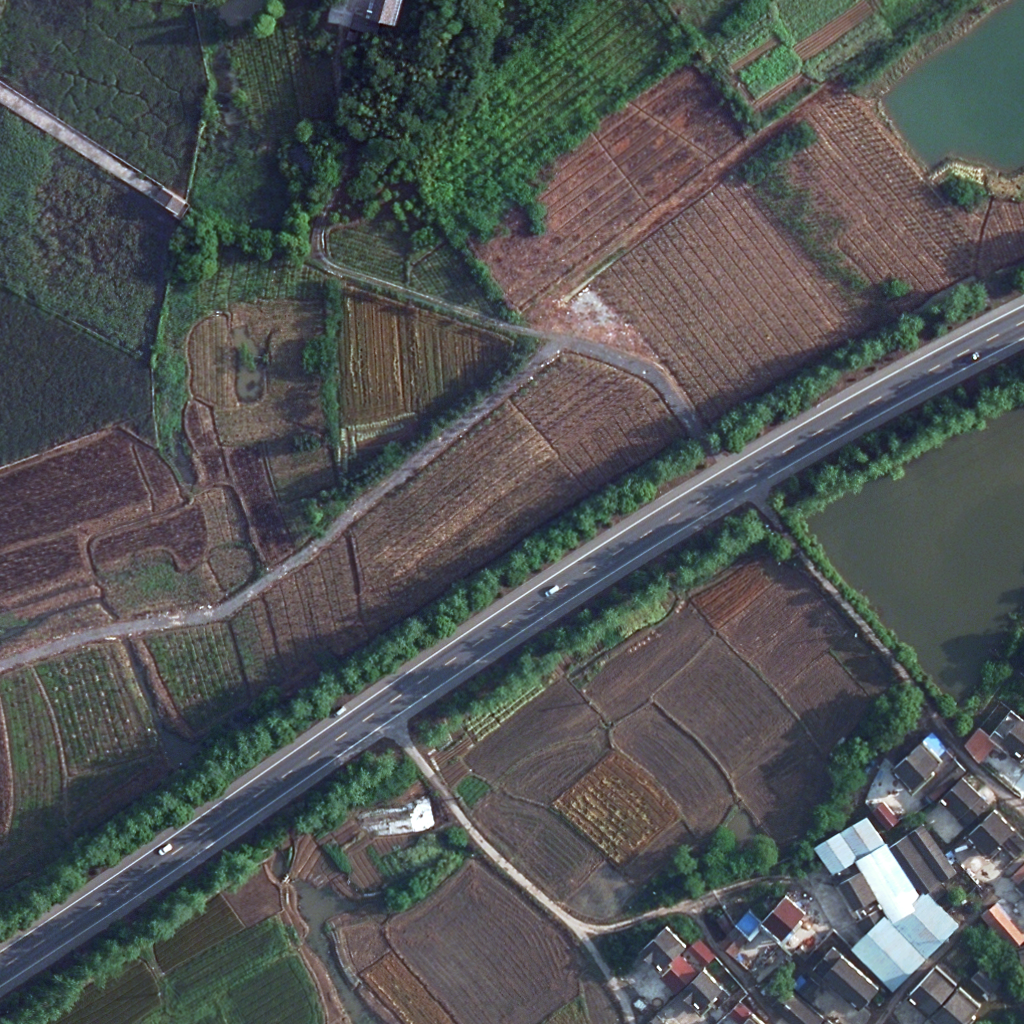}}}&\raisebox{-.5\height}
       {\frame{\includegraphics[width=0.23\textwidth]{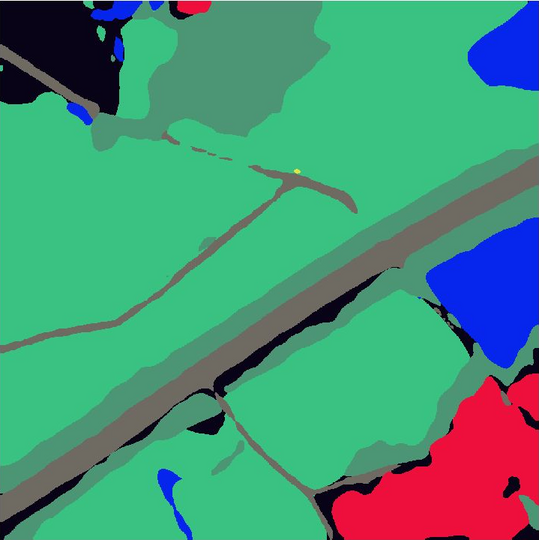}}}\vspace{0.1cm}\\
       \rotatebox[origin=c]{90}{\fontsize{8}{10}\selectfont{Vaihingen}} &\raisebox{-.5\height}{\frame{\includegraphics[width=0.23\textwidth]{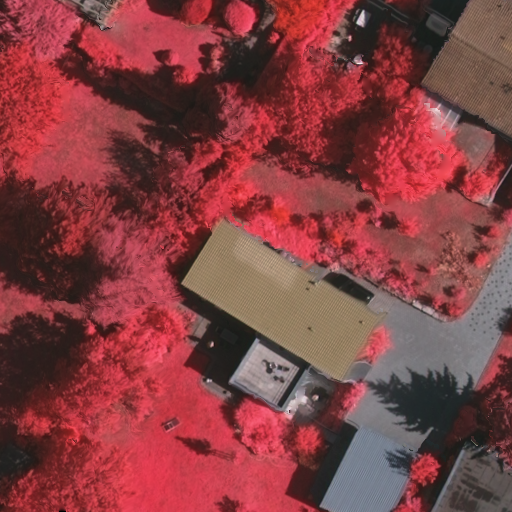}}}&\raisebox{-.5\height}[0pt]{\frame{\includegraphics[width=0.23\textwidth]{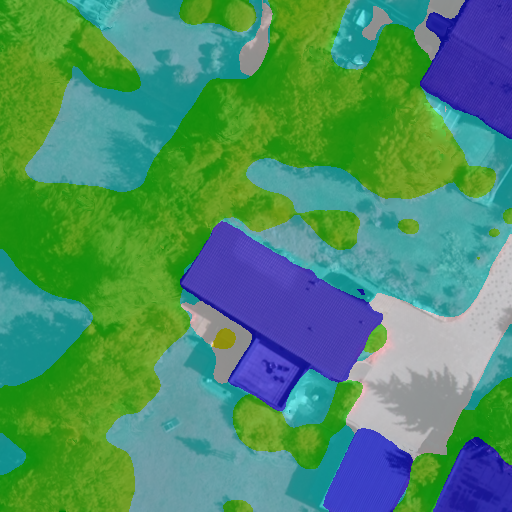}}}&\raisebox{-.5\height}[0pt]{\frame{\includegraphics[width=0.23\textwidth,height=91pt]{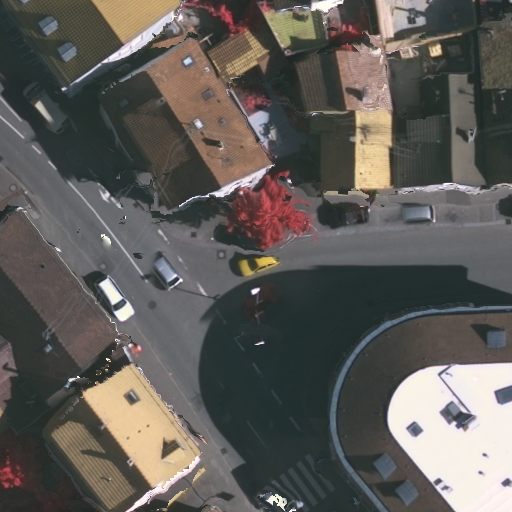}}}&\raisebox{-.5\height}{\frame{\includegraphics[width=0.23\textwidth]{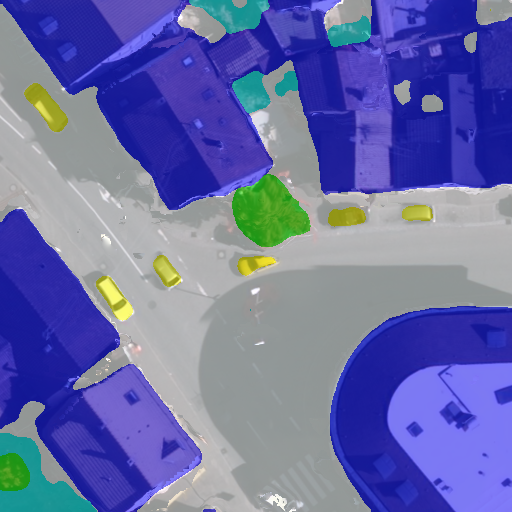}}}\vspace{0.1cm}\\
       \rotatebox[origin=c]{90}{\fontsize{8}{10}\selectfont{GeoNRW}} &\raisebox{-.5\height}{\frame{\includegraphics[width=0.23\textwidth]{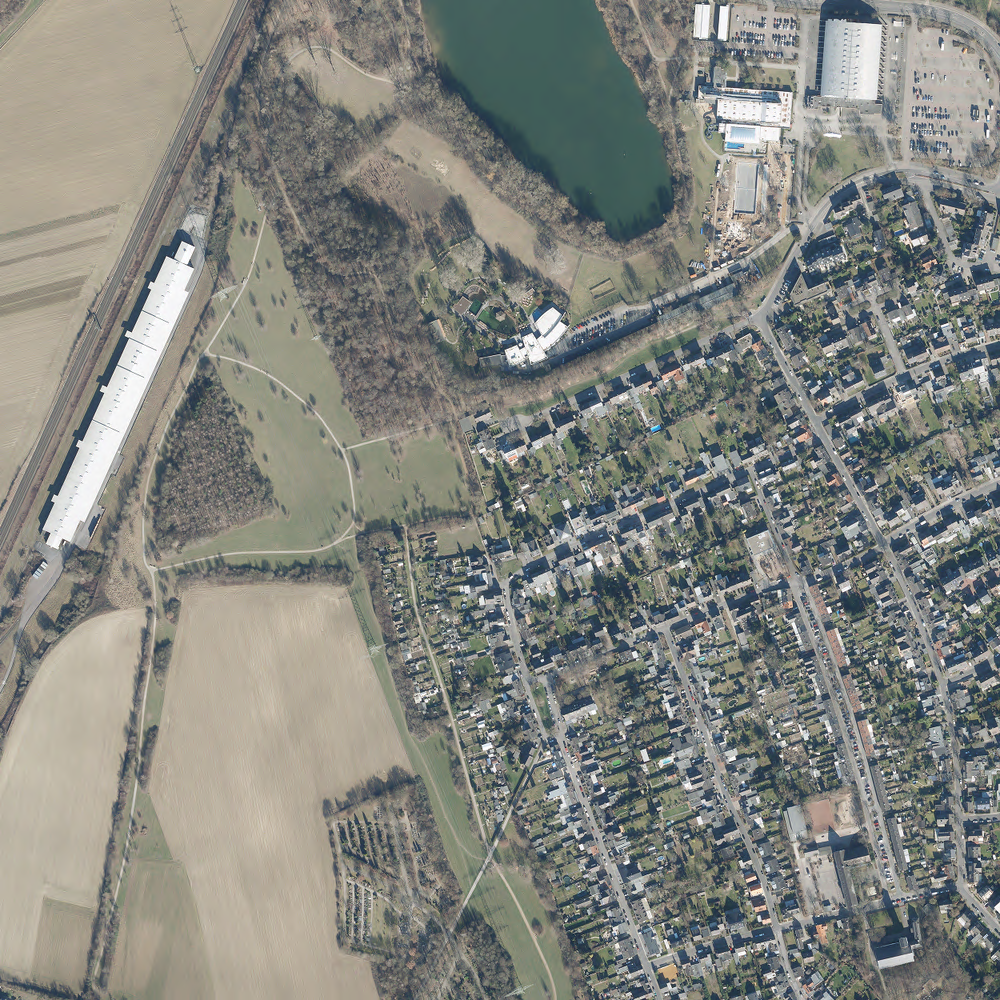}}}&\raisebox{-.5\height}[0pt]{\frame{\includegraphics[width=0.23\textwidth]{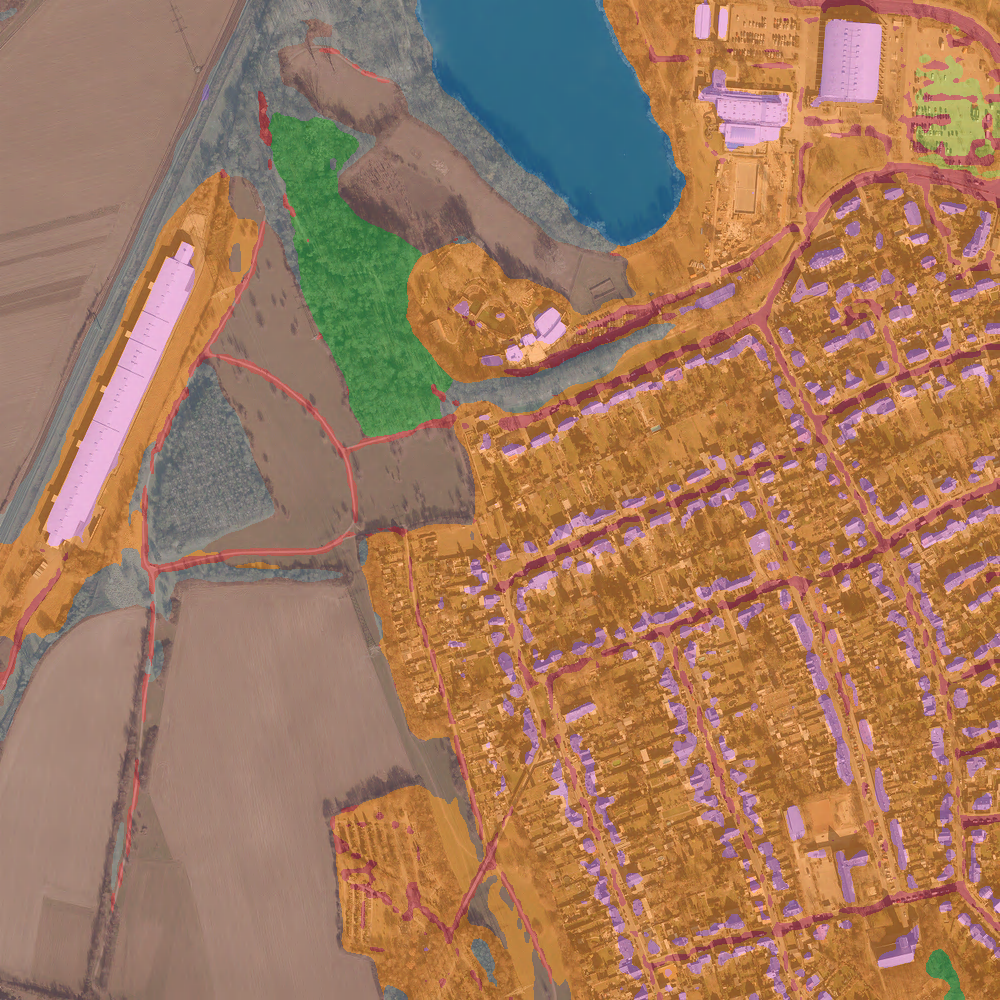}}}&\raisebox{-.5\height}[0pt]{\frame{\includegraphics[width=0.23\textwidth,height=91pt]{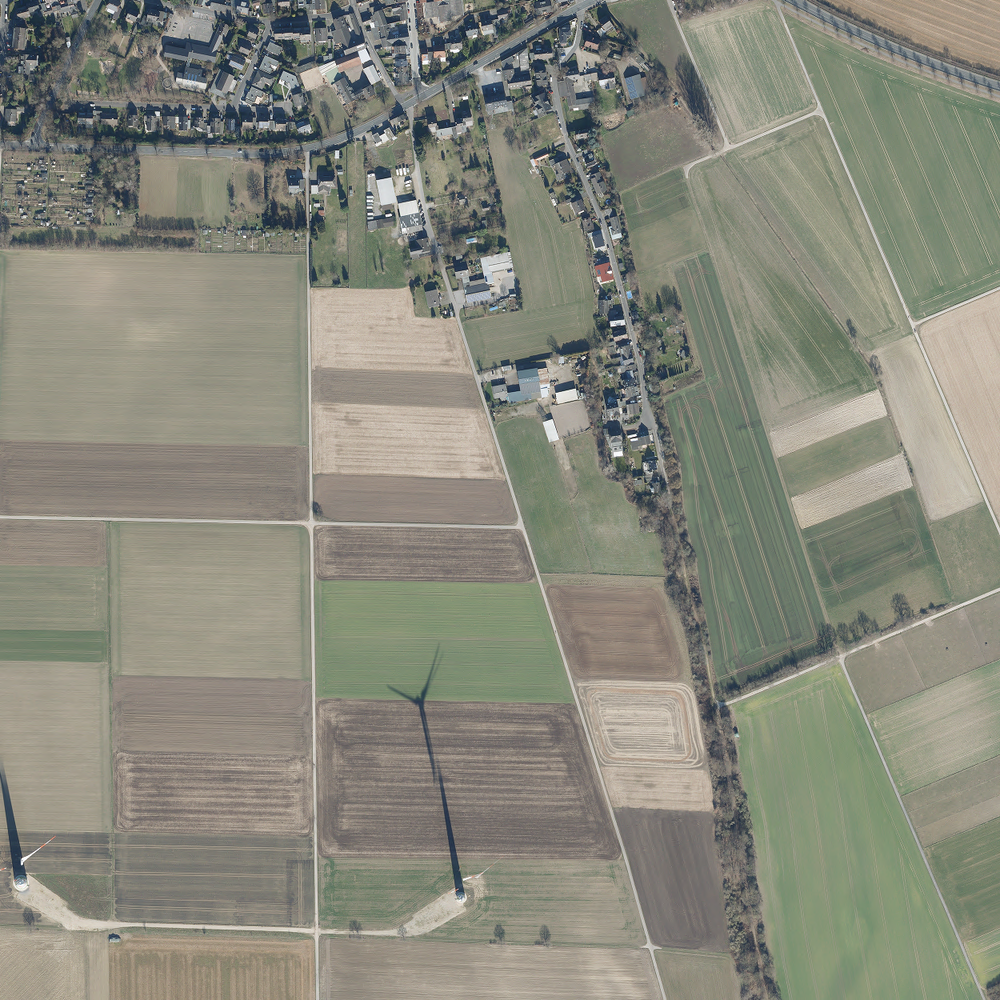}}}&\raisebox{-.5\height}{\frame{\includegraphics[width=0.23\textwidth]{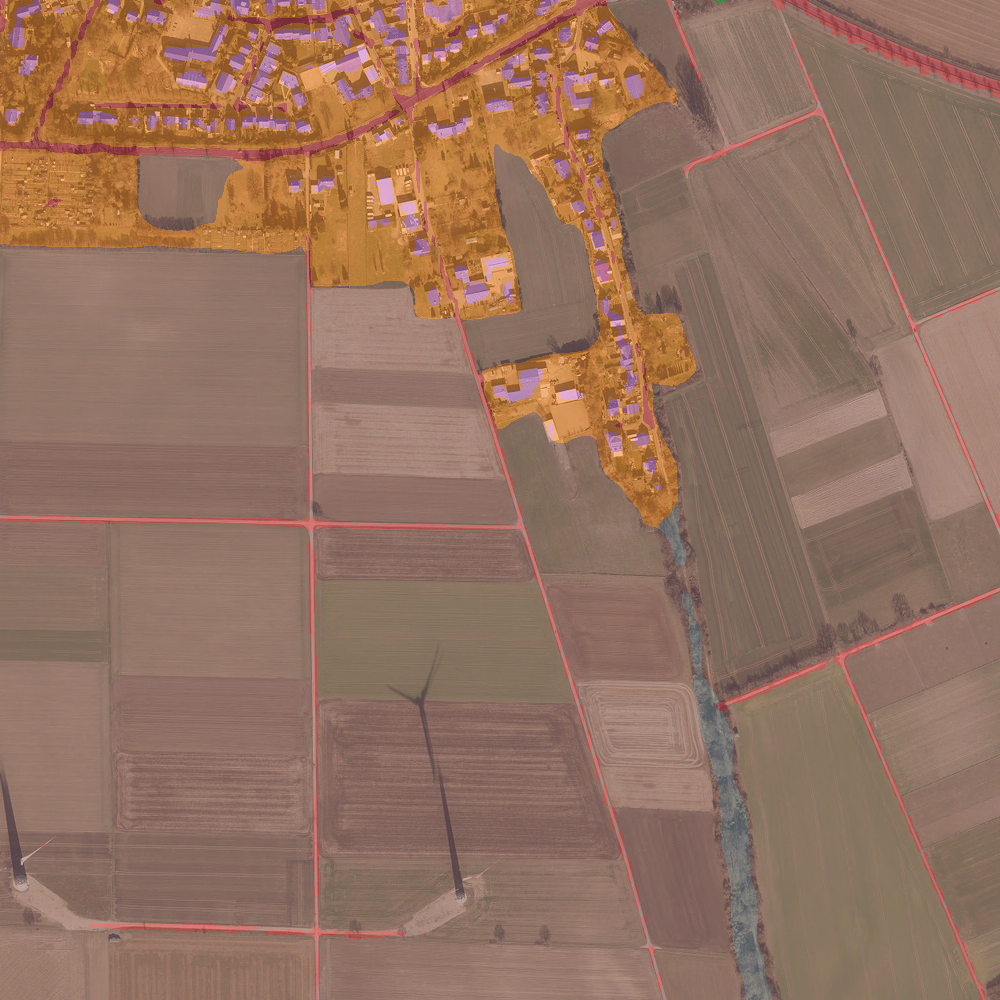}}}\vspace{0.1cm}\\
       \rotatebox[origin=c]{90}{\fontsize{8}{10}\selectfont{SpaceNetv1}} &\raisebox{-.5\height}{\frame{\includegraphics[width=0.23\textwidth]{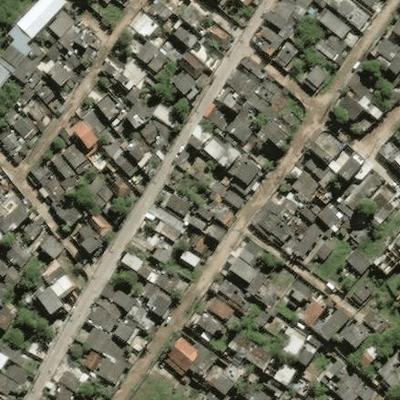}}}&\raisebox{-.5\height}[0pt]{\frame{\includegraphics[width=0.23\textwidth]{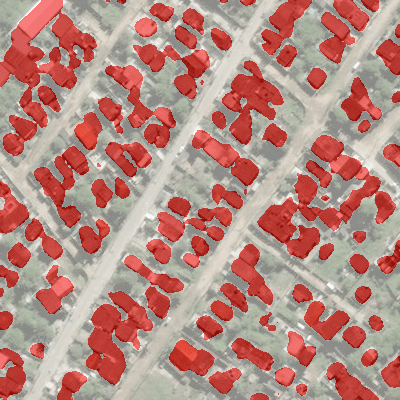}}}&\raisebox{-.5\height}[0pt]{\frame{\includegraphics[width=0.23\textwidth,height=91pt]{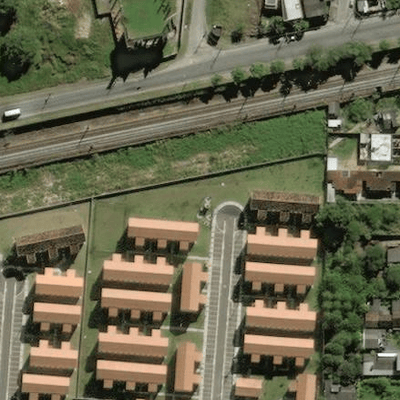}}}&\raisebox{-.5\height}{\frame{\includegraphics[width=0.23\textwidth]{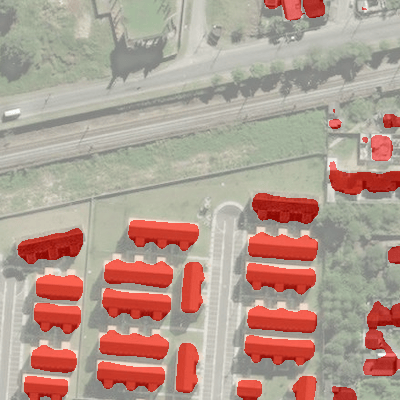}}}
\end{tabular}\caption{\HiResFusedMIM Demonstrates Accurate Segmentation of Buildings and Other Urban Features: Visualized Results on Whu Aerial, LoveDA, Vaihingen, GeoNRW, and SpaceNetv1 datasets.\label{fig:semsegop}}
    \vspace{-0.2cm}
\vspace{-0.2cm}
\end{figure*}

We employed a encoder-decoder architecture with a Swin-Base transformer backbone \cite{swin} for all semantic segmentation tasks.  For Vaihingen and GeoNRW, which provide both RGB and \gls{gl:DSM} data, we designed a DualSwinTransformer backbone, incorporating two separate Swin Transformers - one for RGB input and another for \gls{gl:DSM} input. We chose UPerNet \cite{upernet} from the MMSegmentation library \cite{mmseg} as the architecture of the decoder for all the datasets.

We trained all models for 40k iterations with an input image size of $512\times512$, using the Adam optimizer and the default learning rate of 1e-4. All training settings were consistent with the default configuration in MMSegmentation. 

\begin{table*}[t]
\centering
\caption{ Results for semantic segmentation datasets with RGB modality. \label{tab:semantic_segmentation_results_RGB}}\vspace{0.2cm}
\small
\resizebox{\textwidth}{!}{%
\begin{tabular}{lllllll}
\hline
Method  & Backbone & WHU Aerial & LoveDA & Vaihingen & SpaceNetV1 \\ \hline
SeCo   & ResNet-50 & 86.7  & 43.63 & 68.9  & 73.89    \\
GeoKR   & VGG-16 & -  & - & 74.01  & -  \\
GASSL & ResNet-50 & -   & 48.76 & - & 78.51 \\
SatMAE  & ViT-L & 82.5  & - & 70.6  & 78.07  \\
UperNet(random) & Swin-B    & 88.2  & 44.24 & 67.0 & - &\\
GFM  & Swin-B    & 90.7      & - & \textbf{75.3} & 72.81\\
CMID   & Swin-B  & -  & - & - & 65.98 \\
CtxMIM & Swin-B    & -  & - & - & 79.22 \\
MAE + MTP  & ViT-B + RVSA  & -  & 52.39 & -  & \textbf{79.63} \\
BillionFM  & ViT-L12×4   & -  & 52.38 & - & -\\\hline
\HiResFusedMIM  & Swin-B   & \textbf{91.28}  & \textbf{52.54}& 74.16 & 78.37     \\ \hline

\end{tabular}%
}
\end{table*}

\Cref{tab:semantic_segmentation_results_RGB} presents the quantitative results (mIoU) for \HiResFusedMIM on the RGB-only datasets.  For comparison, we include results from other foundation models and baseline approaches. Note that the results for other models were obtained from their respective publications and are provided here to contextualize the performance of our proposed approach.

\Cref{tab:semantic_segmentation_results_RGB_DSM} compares the performance of three different model configurations on the Vaihingen and GeoNRW datasets: 1) a U-Net with randomly initialized Swin-Base encoder and UPerNet decoder trained in a supervised fashion, 2) the same U-Net architecture but with the encoder initialized using the weights from the \HiResFusedMIM RGB encoder, and 3) a U-Net with the DualSwinTransformer backbone, where both the RGB and \gls{gl:DSM} encoders are initialized with the corresponding weights from \HiResFusedMIMdot.

\HiResFusedMIM demonstrates excellent performance on both the WHU Aerial and LoveDA datasets, achieving \gls{stateoftheart} results and surpassing existing pre-trained models and supervised baselines. On the Vaihingen and SpaceNet datasets, our model achieves mIoU scores that are comparable to other high-performing methods, showcasing its effectiveness across various datasets and for tasks requiring detailed building information. The results on the Vaihingen and GeoNRW datasets, which include \gls{gl:DSM} data, further emphasize the benefits of multi-modal learning. On Vaihingen, initializing the U-Net encoder with weights from the \HiResFusedMIM RGB encoder yields a 74.16\% mIoU, significantly outperforming the randomly initialized model (65.28\% mIoU). Incorporating the \gls{gl:DSM} modality and initializing both encoders with the corresponding \HiResFusedMIM weights results in a further improvement, achieving a 74.4\% mIoU. Similar trends are observed on the GeoNRW dataset, where adding the \gls{gl:DSM} modality leads to a notable boost in performance (61.68\% mIoU compared to 59.39\% with only RGB). These findings highlight the effectiveness of \HiResFusedMIM in leveraging both RGB and \gls{gl:DSM} data to improve segmentation accuracy in complex urban environments.

\begin{table*}[t]
\centering
\caption{ Results for semantic segmentation task datasets with RGB and \gls{gl:DSM} modality. \label{tab:semantic_segmentation_results_RGB_DSM}}\vspace{0.2cm}
\small
\begin{tabular}{llll}
\hline
Method  & Modality  & Vaihingen & GeoNRW \\ \hline
Supervised (Swin-B) & RGB    & 65.28  & 54.8 \\
\HiResFusedMIM    &  RGB    & 74.16  & 59.39 \\ 
\HiResFusedMIM  &  RGB+DEM/DSM   & \textbf{74.4}  & \textbf{61.68} \\\hline
\end{tabular}
\end{table*}

\subsection{Instance Segmentation}

To assess the effectiveness of \HiResFusedMIM for instance segmentation, we conducted experiments using the UBCv2 dataset \cite{ubcv2}, a benchmark designed for building detection and fine-grained classification from very high-resolution (VHR) satellite imagery. The dataset comprises annotated polygons for approximately 0.5 million building instances across 12 roof types, encompassing VHR optical images from 20 diverse cities worldwide, showcasing a wide array of architectural styles and landforms. Seventeen of these cities are also provided with aligned synthetic aperture radar (SAR) imagery, facilitating the development and evaluation of approaches optionally utilizing optical, SAR, or combined modalities. We focused on evaluating the performance of our RGB encoder weights using only the optical imagery from UBCv2.

\begin{figure*}
\tabcolsep=0.05cm
\centering
\begin{tabular}{ccc}
        {\fontsize{8}{10}\selectfont{Ground Truth Image}} & {\fontsize{8}{10}\selectfont{Ground Truth Annotation}} &{\fontsize{8}{10}\selectfont{Segmented Mask}} \\
       \vspace{0.1cm}
       \raisebox{-.5\height}{\frame{\includegraphics[width=0.32\textwidth]{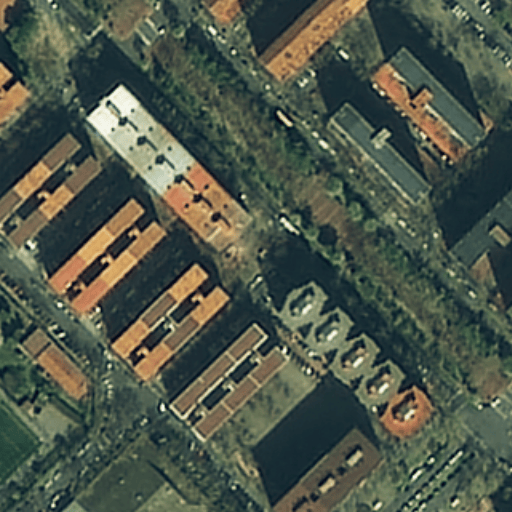}}} & \raisebox{-.5\height}{\frame{\includegraphics[width=0.32\textwidth]{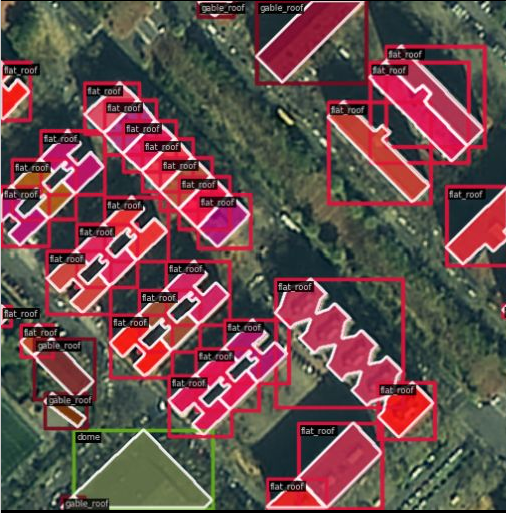}}} & \raisebox{-.5\height}{\frame{\includegraphics[width=0.32\textwidth]{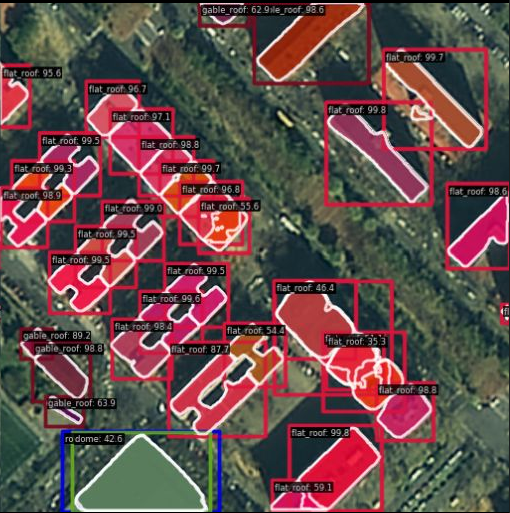}}}\vspace{0.1cm} \\
       \raisebox{-.5\height}{\frame{\includegraphics[width=0.32\textwidth]{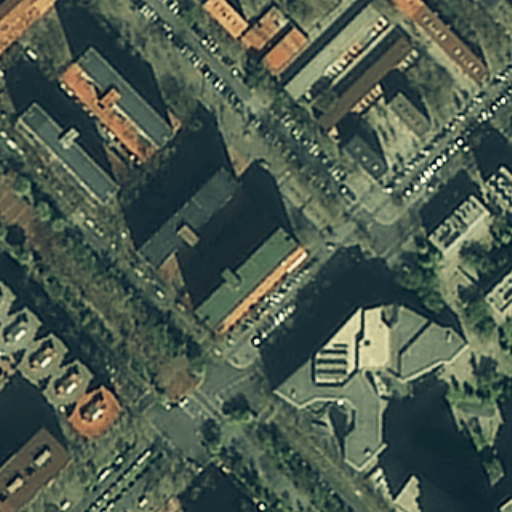}}} & \raisebox{-.5\height}{\frame{\includegraphics[width=0.32\textwidth]{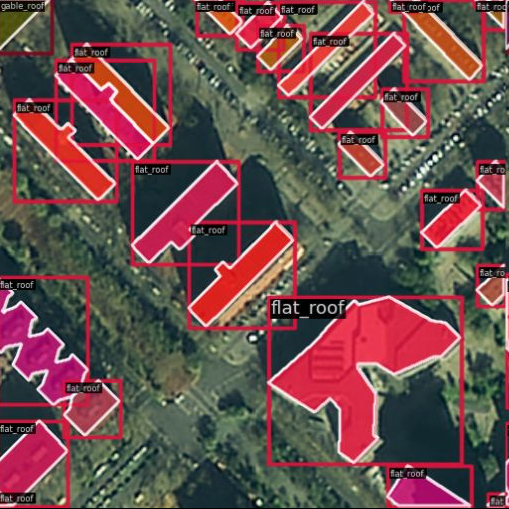}}} & \raisebox{-.5\height}{\frame{\includegraphics[width=0.32\textwidth]{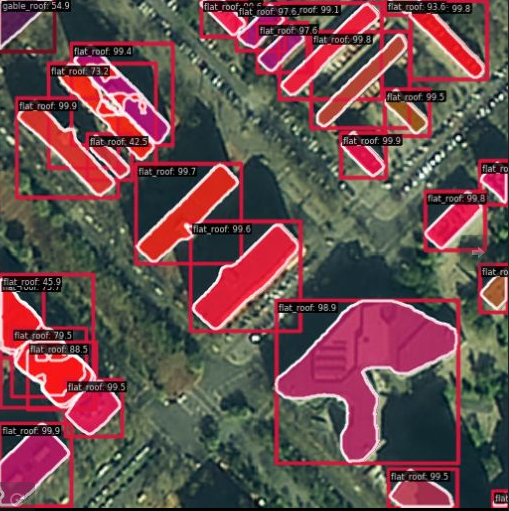}}} \vspace{0.1cm}\\
       \raisebox{-.5\height}{\frame{\includegraphics[width=0.32\textwidth]{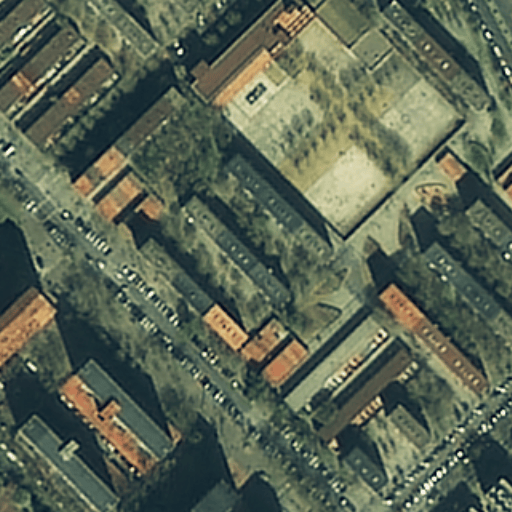}}} & \raisebox{-.5\height}{\frame{\includegraphics[width=0.32\textwidth]{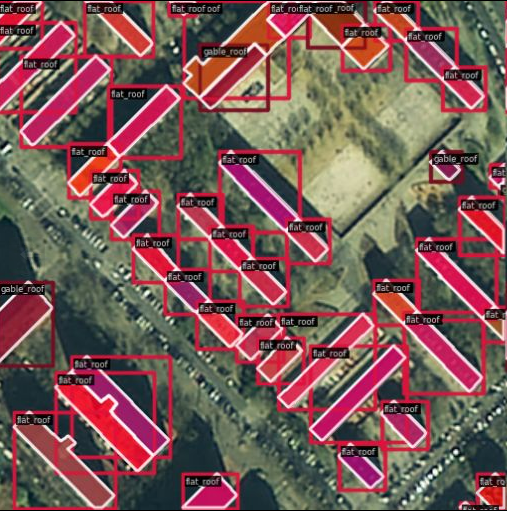}}} & \raisebox{-.5\height}{\frame{\includegraphics[width=0.32\textwidth]{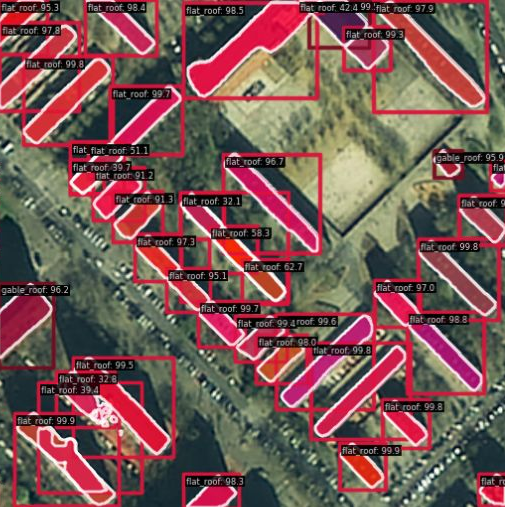}}}\vspace{0.1cm} \\
       \raisebox{-.5\height}{\frame{\includegraphics[width=0.32\textwidth]{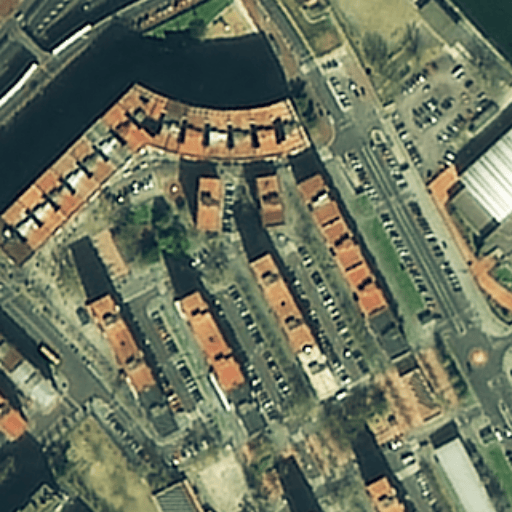}}} & \raisebox{-.5\height}{\frame{\includegraphics[width=0.32\textwidth]{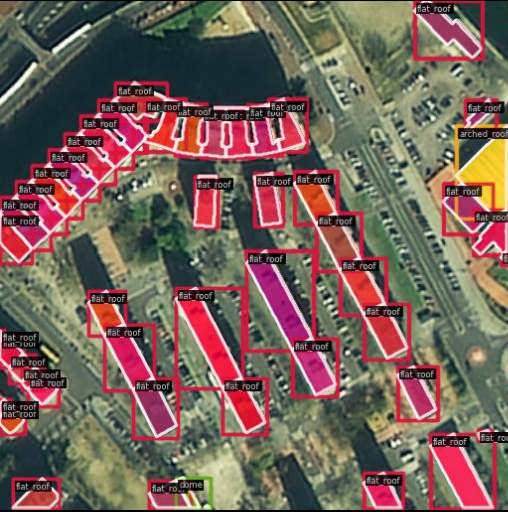}}} & \raisebox{-.5\height}{\frame{\includegraphics[width=0.32\textwidth]{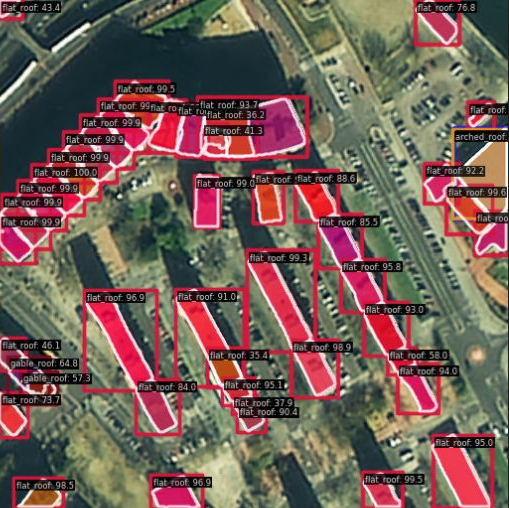}}} \\
\end{tabular}
\caption{\HiResFusedMIM Effectively Segments Individual Buildings in Diverse Urban Environments: Visualized Examples from the UBCv2 Test Set. \label{fig:inssegop}}
\vspace{-0.2cm}
\end{figure*}

We employed the Mask R-CNN model \cite{maskrcnn}, a widely used architecture for instance segmentation, with a Swin Transformer backbone \cite{swin} from the MMDetection library \cite{mmdetection}. We compared our approach to several baselines, including randomly initialized and ImageNet pre-trained Mask R-CNN models, as well as variants utilizing more complex techniques such as Cascade Mask R-CNN with a class-wise geometric transformer (CGT) module introduced in \cite{ubcv2}. These baseline results, obtained from  \cite{ubcv2}, showcase the performance achieved using a ResNet backbone \cite{resnet} for the Mask R-CNN model. In our experiments, we first replaced the ResNet backbone with a randomly initialized Swin-Base \cite{swin} transformer to establish a baseline performance for this backbone. We then initialized the Swin-Base backbone with weights from the RGB encoder of \HiResFusedMIM to evaluate the transferability of the learned representations.

\begin{table*}[t]
\centering
\caption{ Instance segmentation results using AP (\%) on UBCv2 Test set with Fine-Grained roof classes: FL-Flat, GA-Gable, GM-Gambrel, RO- Row, ME- Multiple Eave, H1- Hipped V1, H2- Hipped V2, MA- Mansard, PY- Pyramid, AR-Arched, RE- Revolved and OT-Other.\label{tab:instance_segmentation_results}}\vspace{0.2cm}
\small
\resizebox{\textwidth}{!}{%
\begin{tabular}{llllllllllllllll}
\hline
Method    & AP & $AP_{50}$ & FL & GA & GM & RO & ME & H1 & H2 & MA & PY &AR & RE & OT  \\ \hline
SOLOv2 & 14.2 & 24.0 & 24.6 & 21.3 & 27.7 & 5.8 & 9.5 & 7.6 & 35.0 & 3.6 & 4.6 & 14.8 & 11.2 & 6.1\\
QueryInst & 15.3 & 25.2 & 25.5 & 23.4 & 28.6 & 5.1 & 24.2 & 8.9 & 33.9 & 7.0 & 4.8 & 13.9 & 4.5 &  5.5\\
Mask R-CNN & 15.5 & 25.9 & 25.8 & 24.7 & 29.2 & 5.1 & 12.3 & 10.0 & 39.0 & 5.4 & 5.6 & 16.8 & 5.8 & 7.2\\
Cascade Mask R-CNN & 16.5 & 26.9 & 26.5 & 24.5 & 29.1 & 4.7 & 19.7 & 10.9 & 38.6 & 6.0 & 7.0 & 18.5 & 6.7 & 6.4\\
Mask R-CNN + CGT & 16.3 & 26.3 & 25.2 & 25.1 & 27.4 & 9.0 & 21.6 & 9.8 & 39.6 & 5.7 & 5.8 & 17.2 & 3.2 & 6.8\\
Cascade Mask R-CNN + CGT & 17.1 & 27.1 & 27.2 & 25.0 & 28.6 & 6.6 & 24.9 & 11.1 & 39.8 & 7.7 & 5.8 & 16.4 & 5.7 & 7.2\\
Mask R-CNN (SWIN-B) & 14.8 & 24.5 & 43.8 & 40.8 & 49.4 & 4.3 & 30.2 & 8.7 & 47.2 & 6.7 & 7.7 & 13.3 & \textbf{30.5} & 11.7\\ \hline
\HiResFusedMIM & \textbf{17.7}  & \textbf{28.5} & \textbf{47.32} & \textbf{45.4} & \textbf{54.9} & \textbf{13.7} & \textbf{27.8} & \textbf{16.1} & \textbf{59.0} & \textbf{12.6} & \textbf{12.7} & \textbf{20.9} & 18.6 & \textbf{12.4}  \\ \hline

\end{tabular}%
}
\end{table*}

\Cref{tab:instance_segmentation_results} presents the instance segmentation results on the UBCv2 test set, using Average Precision (AP) and AP$_{50}$ as evaluation metrics. Notably, our experiments revealed that the Mask R-CNN model with a randomly initialized Swin-Base backbone (14.8\% AP) performs less effectively compared to the ResNet-based counterparts reported in the \cite{ubcv2}. However, initializing the Swin-Base backbone with weights from our pre-trained \HiResFusedMIM RGB encoder leads to a substantial improvement, achieving an 17.7\% AP, surpassing all baseline models, including the top-performing Cascade Mask R-CNN + CGT model (17.1\% AP), by a significant margin. This highlights the effectiveness of our pre-training strategy in learning robust and generalizable representations that transfer well to instance segmentation, even when utilizing a different backbone architecture.

\section{Ablation Studies}
\label{sec:ablationstudies}

To gain a deeper understanding of the contributions of different components within \HiResFusedMIMdot, we conducted a series of ablation studies. These studies analyze the impact of the \gls{gl:DSM} encoder and the contrastive loss component.

\subsection{Impact of \gls{gl:DSM} Encoder}
To assess the contribution of the \gls{gl:DSM} encoder, we compared the performance of \HiResFusedMIM with and without the \gls{gl:DSM} modality on datasets that provide both RGB and DSM data (Vaihingen and GeoNRW). As shown in \Cref{tab:semantic_segmentation_results_RGB_DSM}, incorporating the \gls{gl:DSM} data consistently leads to a performance improvement. On the Vaihingen dataset, adding the \gls{gl:DSM} encoder and initializing it with the corresponding \HiResFusedMIM weights increases the mIoU from 74.16\% (RGB-only) to 74.4\%. Similarly, on the GeoNRW dataset, the mIoU increases from 59.39\% (RGB-only) to 61.68\% with the inclusion of the \gls{gl:DSM} modality. These results clearly demonstrate the value of incorporating \gls{gl:DSM} data and training a dedicated DSM encoder for tasks involving building-level analysis. The \gls{gl:DSM} encoder enables the model to capture fine-grained elevation information and to learn richer, multi-modal representations, leading to more accurate segmentation results.

\subsection{Impact of Contrastive Loss}

We investigated the contribution of the contrastive loss component (InfoNCE loss) in the multi-objective loss function used to pre-train \HiResFusedMIMdot. We compared two versions of the model:

\begin{itemize}
    \item \HiResFusedMIM (MIM+Contrastive): The full model, pre-trained with both the reconstruction loss ($\mathcal{L}_{\text{MIM}}$) and the contrastive loss ($\mathcal{L}_{\text{InfoNCE}}$), as described in \cref{sec:methodology}.
    \item \HiResFusedMIM (MIM): A variant of the model trained solely with the reconstruction loss ($\mathcal{L}_{\text{MIM}}$), excluding the contrastive loss.
\end{itemize}
\Cref{tab:contrastive_loss_impact} presents the quantitative results of both models across all downstream tasks.

\begin{table*}[t]
\centering
\caption{Impact of Contrastive Loss on Downstream Task Performance\label{tab:contrastive_loss_impact}}\vspace{0.2cm}
\small
\resizebox{\textwidth}{!}{ % This command rescales the table to fit within the page
\begin{tabular}{lcccccccccccc}
\hline
\multirow{2}{*}{Method} & \multirow{2}{*}{UCM} & \multicolumn{2}{c}{BEN} & \multirow{2}{*}{WHU} & \multirow{2}{*}{LoveDA} & \multicolumn{2}{c}{Vaihingen} & \multicolumn{2}{c}{GeoNRW} & \multirow{2}{*}{SpaceNetV1} & \multicolumn{2}{c}{UBCv2} \\ \cline{3-4} \cline{7-8} \cline{9-10} \cline{12-13}
 &  & 1\% & 10\% &  &  & RGB & RGB+DSM & RGB & RGB+DSM &  & AP & AP$_{50}$ \\ \hline
MIM & 96.7 & 76.1 & 84.68 & 91.13 & 52.35 & 73.84 & 73.1 & 58.9 & 61.13 & 78.32 & 16.9 & 27.4 \\
MIM+Contrastive & \textbf{98.1} & \textbf{76.9} & \textbf{85.1} & \textbf{91.28} & \textbf{52.54} & \textbf{74.16} & \textbf{74.4} & \textbf{59.39} & \textbf{61.68} & \textbf{78.37} & \textbf{17.7} & \textbf{28.5} \\\hline
\end{tabular}
}
\end{table*}

As shown in \Cref{tab:contrastive_loss_impact}, incorporating the contrastive loss during pre-training consistently leads to improved performance across most downstream tasks. While the gains are relatively small on some tasks, we observe more significant improvements on Vaihingen (RGB+\gls{gl:DSM}), GeoNRW (RGB+\gls{gl:DSM}), and UBCv2. This suggests that the contrastive loss is particularly beneficial for tasks that rely heavily on the integration of RGB and \gls{gl:DSM} information. By encouraging the encoders to learn aligned representations for both modalities, the contrastive loss facilitates more effective multi-modal fusion, leading to superior performance on these tasks.

% \subsection{Pre-Training from Scratch vs. Distillation}
% Inspired by the findings of \cite{msgfm}, which investigated the effectiveness of knowledge distillation from ImageNet pre-trained models, we conducted a similar experiment with our RGB encoder. We trained two variants of \HiResFusedMIMdot:
\glsresetall
\section{Conclusion}
\label{sec:Conclusion}
This paper presented \emph{\HiResFusedMIMdot}, a novel pre-trained model specifically designed for building-level analysis in remote sensing. Recognizing the under-utilization of high-resolution \glspl{gl:DSM} in existing pre-trained models, we explored the benefits of integrating \gls{gl:DSM} data with RGB imagery. Our approach involved:

\begin{itemize}
    \item Curating a large-scale, high-resolution dataset: We created a unique dataset of over 368k paired RGB-\gls{gl:DSM} image pairs at a resolution of \SI{0.2}{}-\SI{0.5}{\meter}, meticulously gathered from various sources across Europe.
    \item Designing a dual-encoder SimMIM architecture: Our model utilizes separate encoders for processing RGB and \gls{gl:DSM} data, allowing for specialized feature learning tailored to each modality. This dual-encoder design is further enhanced by incorporating a contrastive loss (InfoNCE) to promote alignment between the learned representations.
    \item Extensive evaluation on diverse downstream tasks: We evaluated \HiResFusedMIM on a wide range of building-related tasks, including classification, semantic segmentation, and instance segmentation, demonstrating its ability to effectively capture and leverage fine-grained, multi-modal information.
\end{itemize}
Our experiments showcase that incorporating \gls{gl:DSM} data during pre-training consistently leads to improved performance, particularly on tasks involving building extraction and segmentation. \HiResFusedMIM achieves best results among its peer models on datasets like WHU Aerial and LoveDA, highlighting its effectiveness in learning robust and generalizable representations for building-focused analysis. Furthermore, our ablation studies confirm the advantages of our dual-encoder architecture and the contribution of the contrastive loss in enhancing multi-modal fusion.

The development of \HiResFusedMIM highlights the potential of leveraging high-resolution \gls{gl:DSM} data for pre-training in remote sensing. By sharing our pre-trained model weights, we aim to facilitate further research and applications in this direction, encouraging the community to explore the benefits of multi-modal learning for building-level analysis.

%\section*{Future Research Directions}
This research has demonstrated the effectiveness of pre-training with high-resolution RGB-\gls{gl:DSM} data for building-level analysis. Building upon these findings, several promising research avenues exist for further exploration. We plan to investigate the application of \HiResFusedMIM to tasks such as \gls{gl:DSM} to DTM generation and building height prediction, leveraging its ability to capture detailed elevation information and learn multi-modal representations. Additionally, we aim to explore the potential of using the strong embeddings from the model within a 3D building reconstruction pipeline, potentially leading to more accurate and efficient reconstruction methods. Further investigation is needed to optimize the fusion techniques and fine-tuning strategies for these tasks, ensuring that the benefits of \HiResFusedMIM are fully realized.

% \appendix

% \section{Sample Appendix Section}
% \label{sec:sample:appendix}

 \bibliographystyle{elsarticle-harv} 
 \bibliography{cas-refs}





% \end{thebibliography}
\end{document}